\documentclass{article}
\usepackage{iclr2027_conference,times}
\usepackage{amsmath,amssymb}
\usepackage{graphicx}
\usepackage{microtype}
\usepackage[hidelinks]{hyperref}
\usepackage[capitalize,noabbrev]{cleveref}
\usepackage{pgfplots}
\usepgfplotslibrary{groupplots}
\pgfplotsset{compat=1.18}
\usepackage[table]{xcolor}
\usepackage{booktabs}
\usepackage{makecell}
\usepackage{etoolbox}

\AtBeginEnvironment{table}{\setlength{\belowcaptionskip}{8pt}}
\AtBeginEnvironment{table*}{\setlength{\belowcaptionskip}{8pt}}
\definecolor{judgmentgray}{HTML}{3F4650}
\definecolor{expressionblue}{HTML}{0072B2}
\definecolor{negorange}{HTML}{D55E00}
\definecolor{stablegreen}{RGB}{0,125,99}

\author{Chaemin Jang, Donggyu Lee, Jeesoo Kim, Dongman Lee$^{\dagger}$, Jihee Kim$^{\dagger}$ \\
Korea Advanced Institute of Science and Technology (KAIST) \\
\texttt{jchaemin@kaist.ac.kr}
}
\iclrfinalcopy

\title{Isolating a Language Model’s Judgment with \\a Label-Free Subspace}

\begin{document}
\maketitle

\begingroup
\renewcommand{\thefootnote}{$\dagger$}
\footnotetext{Co-corresponding authors.}
\endgroup

\begin{abstract}
Language models are often asked to make judgments: whether a proposed answer is correct, whether one statement follows logically from another, or whether a response is safe.
Yet changing the instructions that specify how the judgment should be expressed can make the final answer less reliable, even for the same input.
We ask whether this reflects a change in internal judgment or a failure to express it.
Comparing hidden-state vectors for the same input under different instructions, we find that their differences concentrate in a low-dimensional \emph{expression subspace}.
We identify this subspace without benchmark labels and remove its contribution from a linear predictor to obtain an \emph{isolated judgment}.
Across correctness, entailment, and safety, the isolated judgment remains reliable under instructions not used to identify the subspace, even when the model's final answer and the original predictor degrade.
When the judgment question is negated, isolation raises the predictor's AUROC from 0.504 to 0.827 for entailment and from 0.480 to 0.891 for safety.
We also find that judgment is readable in earlier layers, before the expression subspace begins to disrupt the predictor.
Together, these results provide a way to separate judgment from expression and recover internal judgments even when the model's final answers are unreliable.
\end{abstract}

\section{Introduction}
\label{sec:intro}
Language models are often asked to make judgments: whether a proposed answer is correct, whether one statement follows logically from another, or whether a response is safe.
But making an assessment and expressing it are not necessarily the same.
A model that judges a proposed answer to be correct may be asked to express that judgment as ``Yes,'' ``True,'' a letter, or part of a structured response.
We call the model's internal assessment its \emph{judgment}, and the response it produces its \emph{verdict}.
Different instructions can require different verdicts for the same judgment.
This paper asks whether that judgment remains recoverable when the model fails to express it reliably under a changed instruction.

The verdict alone cannot tell us what failed: did the model change its judgment, or did it fail to express the same judgment in the requested way?
Looking inside the model does not immediately solve the problem either.
The hidden state used to produce the verdict may contain information both about the judgment and about how it should be expressed.
A linear predictor applied to that state, which we call a \emph{readout}, can therefore become less reliable even when the underlying judgment has not changed.

We separate these sources of variation by comparing hidden-state vectors for the same input under different instructions.
Because the input being judged is held fixed, the paired differences capture changes associated with the instruction.
Across inputs, these differences concentrate in a small set of directions.
We use them to identify a low-dimensional \emph{expression subspace} without benchmark labels and remove the readout's component within it to obtain an \emph{isolated judgment}.

On correctness tasks, a linear readout fitted with benchmark labels under the ordinary question ``Is the proposed answer correct?'' reaches 0.815 AUROC.
When the question changes to ``Is the proposed answer incorrect?'', the same readout falls to 0.650, while the isolated judgment reaches 0.802.
Similar recovery occurs when the instructions reverse the response meanings: ``No'' means correct and ``Yes'' means incorrect, or ``False'' means correct and ``True'' means incorrect.
Across correctness, entailment, and safety, the isolated judgment remains reliable under instructions not used to identify the subspace.
When the readout is fitted to the model's own ordinary verdicts, isolation also recovers that assessment even when the verdict under negation expresses the opposite judgment.

Examining the model's layers shows where the readout loses reliability.
Judgment is already readable at intermediate layers, while disruption appears mainly in later layers.
This disruption is concentrated in directions associated with changes in response meaning, such as negating the question or reversing what ``Yes'' and ``No'' mean.
Changing only the response symbols has little effect on the readout.

These findings have implications for both evaluation and the design of internal readouts.
An unreliable verdict need not imply that a useful internal judgment is absent, and a readout can itself be sensitive to how that judgment must be expressed.
Separating these sources of variation provides a way to recover internal judgments despite unreliable verdicts and makes readouts more robust to instruction changes.

Our contributions are:
\begin{itemize}
    \item We show that a model's judgment can remain readable even when its verdict changes.
    \item We identify a low-dimensional \emph{expression subspace} that separates judgment from expression.
    \item We show that the disruption enters late in the network, after the judgment is already readable.
\end{itemize}

\begin{figure*}[t]
    \centering
    \includegraphics[width=\textwidth]{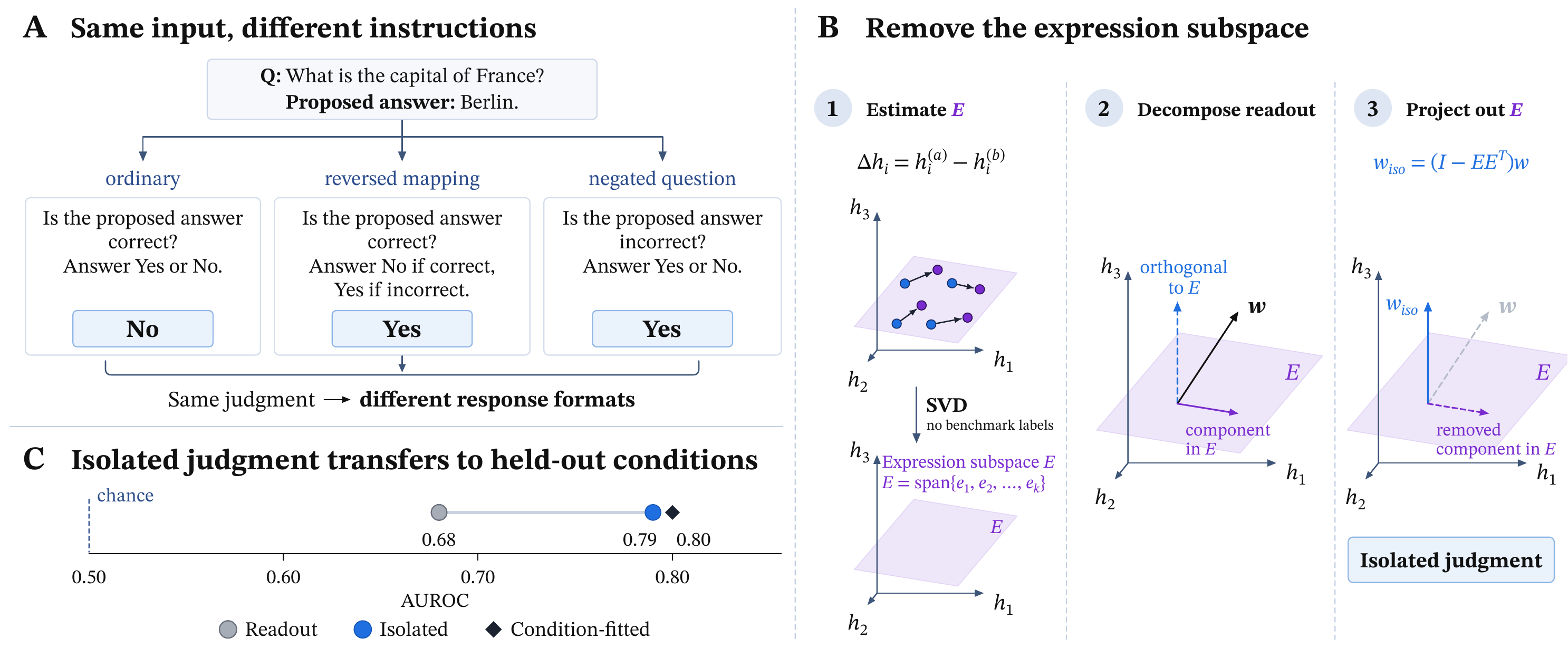}
    \caption{\textbf{Separating judgment from verdict.} \textbf{A:} The same input under three instructions: the verdict changes, the judgment does not. \textbf{B:} Paired differences across instructions span a low-dimensional expression subspace $E$, estimated without labels; removing it from the readout $w$ leaves the isolated judgment. \textbf{C:} On the four held-out conditions, the readout falls to 0.68 AUROC while the isolated judgment reaches 0.79, within 0.01 on average of a readout fitted on each condition.}
    \label{fig:main-overview}
\end{figure*}

\section{Related Work}
\label{sec:related}

\paragraph{Internal representations of truth and judgment.}
A large literature finds that truth, correctness, and factuality are linearly readable from language-model activations \citep{azaria2023internal,marks2024geometry,liu2024hyperplane,burger2024truth,orgad2025know,gekhman2025insideout,servedio2025hidden,han2025factuality}. Models can also evaluate their own candidate answers, and internal states can remain informative when the generated response is misleading or wrong \citep{kadavath2022language,halawi2023overthinking}. At the same time, these signals do not always transfer cleanly across datasets, tasks, models, or logical transformations \citep{levinstein2024still,bao2025probing,poulis2026limits,cheang2026recall}. Most closely related, \citet{burger2024truth} separate a general truth direction from a polarity-sensitive component under affirmation and negation. We instead study the judgment a model forms for a particular input being judged and ask whether that judgment can be separated from the way it must be expressed.

\paragraph{From deciding to expressing a response.}
Model behavior is highly sensitive to prompt format, answer priors, demonstration order, label mappings, and response symbols \citep{zhao2021calibrate,lu2022fantastically,webson2022prompts,min2022rethinking,sclar2024quantifying,robinson2023leveraging,zheng2024large,wei2023symbol}. Mechanistic work increasingly suggests that selecting an answer and mapping it to the required response are distinct computations \citep{lieberum2023does,wiegreffe2025answer,wong2026bind}. Related work has also identified activation directions controlling instruction-specific properties such as output format and stage-like separation between content and surface form \citep{stolfo2025instruction,wendler2024llamas}. Our results identify the corresponding structure for judgments: the judgment is readable before the expression subspace begins to affect the readout, and paired changes in the instruction reveal the subspace itself.

\paragraph{Probing, representation removal, and intervention.}
Decodability alone does not establish how a representation is used, motivating controls and counterfactual interventions in probing work \citep{hewitt2019designing,elazar2021amnesic,park2024linear}. Unsupervised methods can recover latent directions from paired or logically related examples, although such objectives need not identify the intended latent variable \citep{burns2023latent,farquhar2023challenges}. Other work removes linearly represented concepts or steers behavior along activation directions \citep{ravfogel2020inlp,belrose2023leace,subramani2022steering,li2023inference,zou2023representation}. Our estimator instead uses paired states in which the input being judged is fixed and only the instruction changes. The resulting expression subspace requires no benchmark labels to estimate and can be removed to isolate a judgment that transfers to conditions held out from its estimate.
\section{Setup}
\label{sec:setup}

\paragraph{Experimental setting.}
We study three binary judgments: whether a proposed answer is correct, whether a hypothesis is entailed, and whether a response is safe. Our main experiments use five instruction-tuned models across six datasets; we additionally replicate the core result prospectively on five further models up to 32B. The fourteen primary model--dataset cells in the main tables come from the first four models on the four correctness datasets; Gemma-2-9B is the prospective fifth model and is reported separately (Appendix~\ref{app:heldout}).

\begin{table}[t]
\caption{Experimental setting. For each listed pair, the first response expresses a positive judgment. Held-out conditions are not used to identify the subspace and are introduced only when testing whether the judgment can be recovered under conditions held out from the estimate.}
\centering
\begin{tabular}{p{0.20\linewidth}p{0.74\linewidth}}
\toprule
\textbf{Component} & \textbf{Setting} \\
\midrule
Judgments
& Correctness, entailment, safety \\

Datasets
& GSM8K, MMLU, ARC-Challenge, TriviaQA, ANLI, BeaverTails \\

Models
& Qwen2.5-7B, Llama-3.1-8B, Mistral-7B, Qwen2.5-14B, Gemma-2-9B \\

Response formats
& Yes/No, True/False, A/B, B/A, 1/0, Correct/Incorrect, sentence \\

Held-out conditions
& No/Yes, False/True, 0/1, negated question \\
\bottomrule
\end{tabular}

\label{tab:setup}
\end{table}

For correctness, each model first generates a candidate answer and is then asked whether that answer is correct. For entailment, we use ANLI and ask whether the hypothesis follows from the premise; for safety, we use BeaverTails and ask whether the response is safe. In every case, we keep the input being judged fixed and change only the instruction used to elicit the judgment.

\paragraph{Prompt structure.}
For correctness, the prompt consists of the question, the candidate answer, and the instruction. For example, under the Yes/No response format, the prompt has the following form:

\begin{quote}
\small
\texttt{[question]}\\[0.5em]
\texttt{[candidate answer]}\\[0.5em]
\texttt{Is the proposed answer correct? Answer Yes or No.}
\end{quote}

This ordering lets us read the hidden state before the judgment question, after the judgment question, and immediately before the verdict. In one control, we specify the response format before the judgment question to test whether the judgment is already readable before verdict generation.

\noindent\begin{minipage}{\linewidth}
The instructions for several conditions are:

\begin{center}
\small
\renewcommand{\arraystretch}{2}
\begin{tabular}{p{0.19\linewidth} p{0.72\linewidth}}
\textbf{Condition} & \textbf{Instruction} \\ \hline

Yes/No &
\texttt{Is the proposed answer correct? Answer Yes or No.} \\

A/B &
\makecell[l]{\texttt{Is the proposed answer correct?}\\
\texttt{Answer A for correct or B for incorrect.}} \\

Negated question &
\texttt{Is the proposed answer incorrect? Answer Yes or No.} \\

\end{tabular}
\renewcommand{\arraystretch}{1}
\end{center}
\end{minipage}

\paragraph{Readout.}
We read the model's final-layer hidden state immediately before it produces its verdict and fit a simple linear readout to distinguish positive from negative benchmark labels. After standardizing the hidden states, the readout direction is the difference between the mean state for positive and negative examples; we use this class mean rather than a tuned probe because tuned probes transfer worse across formats (Appendix~\ref{app:transfer}). Unless stated otherwise, we fit the readout under the Yes/No condition. When applying it to another condition or dataset, we keep its direction and scaling fixed; no benchmark labels from the target condition are used.

\paragraph{Evaluation.}
We divide questions into five disjoint groups, fitting the readout on four and evaluating it on the fifth in turn. All conditions and wordings of the same underlying question are kept in the same group, ensuring that no question appears in both training and evaluation. We report AUROC, which measures how well the readout separates positive from negative benchmark labels without requiring a fixed decision threshold. Uncertainty intervals are estimated by repeatedly resampling questions within each model--dataset pair and resampling across pairs for aggregate results.

\section{The Readout Is Linked to the Model's Judgment}
\label{sec:readout}

The readout becomes more informative after the model is asked to judge the candidate, and intervening along its direction changes the verdict the model produces. Its AUROC rises from 0.741 before the judgment question to 0.812 after the question and 0.818 immediately before the verdict. Moreover, intervening along the readout direction systematically shifts the model toward the corresponding verdict. Together, these results connect the readout to the model’s judgment process and provide a starting point for separating the judgment from how it is expressed.

\paragraph{The readout becomes more informative when the model judges the candidate.}
At the final token of the candidate answer, before any judgment question is asked, the readout reaches 0.741 AUROC. After the judgment question, it rises to 0.812 and reaches 0.818 immediately before the verdict, with post-judgment AUROC exceeding pre-judgment AUROC in all fourteen model–dataset pairs. Thus, the signal captured by the readout becomes more predictive once the model is asked to evaluate the candidate, beyond what is already readable from the candidate answer alone. When the response format is specified before the judgment question, the same pattern remains: the readout reaches 0.808 AUROC after the judgment question and 0.823 immediately before the verdict. Full prompt-position results and per-cell values are reported in Appendix~\ref{app:position}. 

The readout also does not depend on who wrote the candidate: fitted on the judge's own answers and applied to another model's answers to the same questions, it is within 0.009 of a readout fitted on those states, and it reads dataset-written candidates at 0.85--0.89 (Appendix~\ref{app:candidates}).
\paragraph{Intervening along the readout direction shifts the verdict.}
Adding the readout direction at an intermediate layer systematically shifts the model toward the verdict corresponding to a positive judgment. The logit difference between the responses meaning “correct” and “incorrect” increases by 2.39 logits per unit under Yes/No, 1.30 under A/B, and 0.46 under B/A. This effect appears in all 14 model–dataset pairs under Yes/No and A/B and in 13 of 14 under B/A, while random directions and directions fitted to shuffled benchmark labels have essentially no effect. Thus, the readout direction is not merely predictive of the verdict: intervening along it systematically changes the verdict the model produces (Appendix~\ref{app:steering}). Table~\ref{tab:readout-evidence} summarizes these results.

\begin{table}[t]
\caption{
\textbf{The readout is linked to the model's judgment.}
The readout becomes stronger after the model is asked to judge the candidate answer,
and intervening along its direction changes the model's verdict.
}
\centering
\small

\begin{minipage}[t]{0.42\linewidth}
\centering
\textbf{Readout across prompt position}

\vspace{3pt}

\begin{tabular}{lc}
\toprule
Position & AUROC \\
\midrule
Before judgment question & 0.741 \\
After judgment question  & 0.812 \\
Before verdict            & 0.818 \\
\bottomrule
\end{tabular}
\end{minipage}
\hfill
\begin{minipage}[t]{0.54\linewidth}
\centering
\textbf{Intervention along the readout direction}

\vspace{3pt}

\begin{tabular}{lcc}
\toprule
Response format & Logit change / unit & Positive cells \\
\midrule
Yes/No & 2.39 & 14/14 \\
A/B    & 1.30 & 14/14 \\
B/A    & 0.46 & 13/14 \\
\bottomrule
\end{tabular}
\end{minipage}

\label{tab:readout-evidence}
\end{table}

\section{A Low-Dimensional Expression Subspace Separates Judgment from Expression}
\label{sec:report}

Across changes in response form, the readout transfers nearly unchanged: fitted under Yes/No, it is within 0.009 AUROC on average of a readout fitted directly on A/B, B/A, or sentence responses (Appendix~\ref{app:transfer}). By contrast, the readout degrades when the instruction changes what the response means, even though the content being judged is fixed. The corresponding changes in the hidden state are concentrated in a small set of directions, which we call the \emph{expression subspace}. Removing this subspace from the readout yields the \emph{isolated judgment}, which remains reliable under held-out changes in how the judgment must be expressed.

\subsection{Identifying the Expression Subspace}
\label{sec:expression-subspace}

The expression subspace captures directions that change consistently when the same input is judged under different instructions. For each input, we take the difference between its hidden states under two instructions and identify the directions that recur across these paired differences:

\[
\underbrace{h_i^{(a)} - h_i^{(b)}}_{\substack{\text{same input}\\
                                               \text{different instructions}}}
\quad\xrightarrow{\ \text{across inputs}\ }\quad
\underbrace{E}_{\text{expression subspace}}
\quad\xrightarrow{\ \text{remove from readout}\ }\quad
\underbrace{(I-EE^\top)w}_{\text{isolated judgment}} .
\]

Here, the columns of $E$ are the leading directions of the centered paired
hidden-state differences, obtained by SVD. We select the rank by split-half
stability; across cells, the selected rank has a median of eight.

The expression subspace is estimated from paired states under five response
forms: Yes/No, True/False, A/B, B/A, and sentence responses, together with
affirmative and negated judgment pairs from entailment and safety. No benchmark
labels are used to estimate the subspace. The conditions in which No, False,
or 0 means correct, as well as the negated correctness question, are excluded
from estimation and used only for evaluation. All subspaces are estimated on
training questions and re-estimated within each fold. Full details are given
in Appendix~\ref{app:estimation}.

\subsection{The Isolated Judgment Transfers to Held-Out Conditions}
\label{sec:heldout-main}

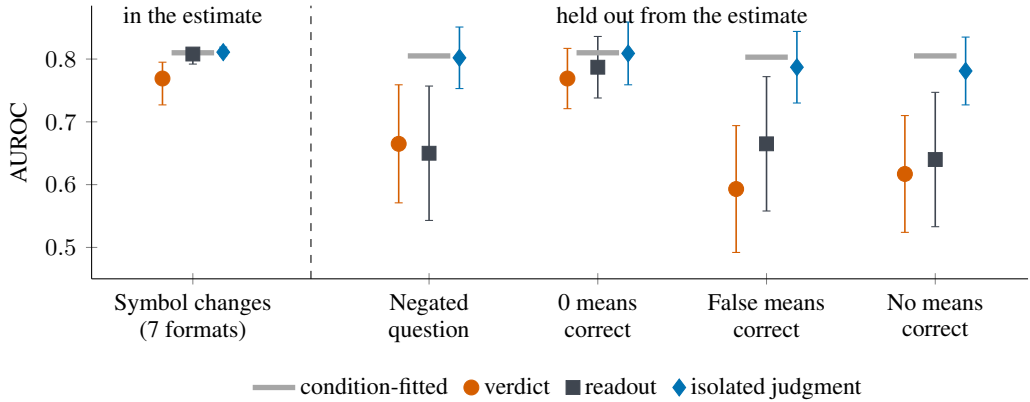
\begin{figure}[t]
\centering
\begin{tikzpicture}
\begin{axis}[
  width=\textwidth, height=5.2cm,
  ymin=0.45, ymax=0.885, ytick={0.5,0.6,0.7,0.8}, ylabel={AUROC},
  xmin=-0.6, xmax=4.95,
  xtick={0,1.4,2.4,3.4,4.4},
  xticklabels={{Symbol changes\\(7 formats)},{Negated\\question},{0 means\\correct},{False means\\correct},{No means\\correct}},
  xticklabel style={align=center, font=\footnotesize},
  tick label style={font=\footnotesize}, label style={font=\footnotesize},
  axis lines=left, axis line style={-},
  legend style={at={(0.5,-0.32)}, anchor=north, legend columns=4, draw=none, font=\footnotesize, /tikz/every even column/.append style={column sep=0.4em}},
  error bars/y dir=both, error bars/y explicit,
  error bars/error bar style={line width=0.6pt}, error bars/error mark options={rotate=90, mark size=1.5pt},
  clip=false,
]
\addplot[only marks, mark=-, mark size=8pt, line width=2pt, color=gray!70] coordinates {(0,0.810) (1.4,0.805) (2.4,0.810) (3.4,0.803) (4.4,0.805)};
\addlegendentry{condition-fitted}
\addplot[only marks, mark=*, mark size=2.8pt, color=negorange] coordinates {
  (-0.18,0.769) += (0,0.026) -= (0,0.042) (1.22,0.665) +- (0,0.094) (2.22,0.769) +- (0,0.048) (3.22,0.593) +- (0,0.101) (4.22,0.617) +- (0,0.093)};
\addlegendentry{verdict}
\addplot[only marks, mark=square*, mark size=2.6pt, color=judgmentgray] coordinates {
  (0,0.808) += (0,0.008) -= (0,0.016) (1.4,0.650) +- (0,0.107) (2.4,0.787) +- (0,0.049) (3.4,0.665) +- (0,0.107) (4.4,0.640) +- (0,0.107)};
\addlegendentry{readout}
\addplot[only marks, mark=diamond*, mark size=3.2pt, color=expressionblue] coordinates {
  (0.18,0.811) += (0,0.008) -= (0,0.006) (1.58,0.802) +- (0,0.049) (2.58,0.809) +- (0,0.050) (3.58,0.787) +- (0,0.057) (4.58,0.781) +- (0,0.054)};
\addlegendentry{isolated judgment}
\draw[dashed] (axis cs:0.7,0.45) -- (axis cs:0.7,0.885);
\node[font=\footnotesize] at (axis cs:0,0.87) {in the estimate};
\node[font=\footnotesize] at (axis cs:2.9,0.87) {held out from the estimate};
\end{axis}
\end{tikzpicture}
\caption{Across conditions, the readout can degrade while the isolated judgment remains close to a readout fitted directly on the target condition. Left: the seven response formats used to identify the expression subspace, averaged, with whiskers spanning the seven format means. Right: the four held-out conditions, with 95\% intervals over the fourteen model--dataset cells. Gray ticks mark the condition-fitted readout. Means over fourteen cells and five instruction wordings; the sentence format has no native verdict margin.}
\label{fig:conditions}
\end{figure}

\paragraph{Held-out conditions.}
Under instruction changes excluded from subspace estimation, the readout degrades while the isolated judgment remains close to a readout fitted directly on each target condition. Under the negated correctness question, the readout falls to 0.650 AUROC, while the isolated judgment reaches 0.802, compared with 0.805 for the condition-fitted readout (Appendix~\ref{app:negation}). When No, False, or 0 means correct, the readout reaches 0.640, 0.665, and 0.787, while the isolated judgment reaches 0.781, 0.787, and 0.809, respectively. Figure~\ref{fig:conditions} shows the full pattern across conditions, and Table~\ref{tab:heldout-main} reports the four designated held-out conditions.

\begin{table}[t]
\caption{Performance on the four held-out conditions. Values are AUROC against benchmark labels, averaged over fourteen model--dataset cells and five instruction wordings. The isolated judgment remains close to a readout fitted directly on each condition.}
\centering
\small
\begin{tabular}{lcccc}
\toprule
Condition & Verdict & Readout & Isolated & Condition-fitted \\
\midrule
No means correct    & 0.617 & 0.640 & 0.781 & 0.805 \\
False means correct & 0.593 & 0.665 & 0.787 & 0.803 \\
0 means correct     & 0.769 & 0.787 & 0.809 & 0.810 \\
Negated question    & 0.665 & 0.650 & 0.802 & 0.805 \\
\bottomrule
\end{tabular}

\label{tab:heldout-main}
\end{table}

\paragraph{Other judgment tasks.}
The same pattern extends to entailment and safety. For these tasks, the expression
subspace is estimated from paired differences among each task's five response forms,
together with negation pairs from correctness, leaving the negated entailment and
safety questions held out. Under negation, the readout falls to 0.504 AUROC on ANLI
and 0.480 on BeaverTails, while the isolated judgment reaches 0.827 and 0.891,
respectively, nearly matching readouts fitted directly on those conditions
(0.827 and 0.881; Appendix~\ref{app:tasks}).

\paragraph{Comparison with supervised isolation.}
A supervised class-mean baseline using labeled states under both A/B and B/A reaches similar performance on the held-out conditions (Appendix~\ref{app:heldout}). The advantage of the expression subspace is therefore not higher held-out AUROC, but that it identifies the directions associated with expression from paired states alone, without requiring labels from a second response form.

\paragraph{Additional models.}
The same pattern extends beyond the models used to develop the method. Gemma-2-9B-IT
was added after the method was fixed, with predictions specified in advance, and
shows the same result (Appendix~\ref{app:heldout}). In Qwen2.5-32B, the readout
falls to 0.42 AUROC under the negated question, while the isolated judgment comes
within 0.007 of a readout fitted directly on that condition. Across five additional
models spanning 7B to 32B, the isolated judgment is within 0.02 of the
condition-fitted readout in every model (Appendix~\ref{app:models}).

\subsection{Robustness and Alternative Explanations}

\paragraph{Alternative explanations.}
The expression subspace is not simply capturing generic differences between conditions. Removing directions that best encode condition identity, the leading directions of overall hidden-state variation, or a random subspace of the same rank does not recover the readout on the held-out conditions. Moreover, the same pattern remains when the expression subspace is estimated only from pairs on which the model expresses the same judgment under both instructions. Thus, the subspace does not arise simply from examples in which changing the instruction changes the model's judgment (Appendix~\ref{app:controls}).

\paragraph{Robustness across estimation settings.}
The expression subspace is not tied to the particular prompts used to estimate it. A subspace estimated from one set of prompt wordings remains effective on unseen wordings, and a subspace estimated from the original instruction family transfers to a different instruction family. These results suggest that the expression subspace captures structure that recurs across prompts, rather than a property specific to the wordings or instruction family used for estimation (Appendix~\ref{app:controls}).

\section{The Judgment Remains Readable When the Verdict Changes}
\label{sec:isolate}
A model can change its verdict under a new instruction while preserving the judgment
that remains readable internally. Under the negated question and reversed response
mappings, the verdict and readout degrade substantially while the isolated judgment
changes little. More directly, when negation causes the model to give different
verdicts for the same input, the isolated judgment remains aligned with the judgment
expressed under the affirmative question. The changed verdict therefore reflects a
change in expression rather than a loss of the judgment that was already present.

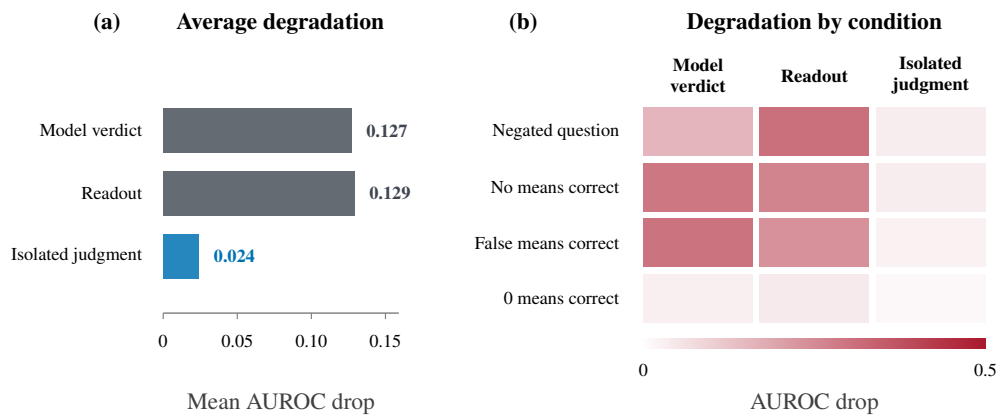
\begin{figure}[t]
\centering

\definecolor{heatred}{RGB}{170,25,45}

\resizebox{0.95\linewidth}{!}{%
\begin{tikzpicture}[x=1cm,y=1cm]


\node[
    anchor=east,
    font=\small\bfseries
] at (-0.45,5.35) {(a)};

\node[
    anchor=center,
    font=\small\bfseries
] at (1.58,5.35) {Average degradation};

\node[
    anchor=east,
    font=\small\bfseries
] at (5.18,5.35) {(b)};

\node[
    anchor=center,
    font=\small\bfseries
] at (8.79,5.35) {Degradation by condition};


\node[anchor=east,font=\scriptsize]
    at (-0.16,3.90) {Model verdict};

\node[anchor=east,font=\scriptsize]
    at (-0.16,3.05) {Readout};

\node[anchor=east,font=\scriptsize]
    at (-0.16,2.20) {Isolated judgment};

\fill[judgmentgray!80]
    (0,3.60) rectangle (2.54,4.20);

\fill[judgmentgray!80]
    (0,2.75) rectangle (2.58,3.35);

\fill[expressionblue!85]
    (0,1.90) rectangle (0.48,2.50);

\node[anchor=west,font=\scriptsize\bfseries,text=judgmentgray]
    at (2.62,3.90) {0.127};

\node[anchor=west,font=\scriptsize\bfseries,text=judgmentgray]
    at (2.66,3.05) {0.129};

\node[anchor=west,font=\scriptsize\bfseries,text=expressionblue]
    at (0.56,2.20) {0.024};

\draw[
    black!50,
    line width=0.55pt
] (0,1.43) -- (3.18,1.43);

\foreach \x/\lab in {
    0/0,
    1.00/{0.05},
    2.00/{0.10},
    3.00/{0.15}
}{
    \draw[
        black!50,
        line width=0.45pt
    ] (\x,1.43) -- (\x,1.35);

    \node[
        anchor=north,
        font=\scriptsize
    ] at (\x,1.27) {\lab};
}


\node[
    align=center,
    font=\scriptsize\bfseries,
    text width=1.15cm
] at (7.22,4.64)
{Model\\verdict};

\node[
    align=center,
    font=\scriptsize\bfseries,
    text width=1.15cm
] at (8.79,4.64)
{Readout};

\node[
    align=center,
    font=\scriptsize\bfseries,
    text width=1.25cm
] at (10.36,4.64)
{Isolated\\judgment};

\node[anchor=east,font=\scriptsize]
    at (6.28,3.88) {Negated question};

\node[anchor=east,font=\scriptsize]
    at (6.28,3.13) {No means correct};

\node[anchor=east,font=\scriptsize]
    at (6.28,2.38) {False means correct};

\node[anchor=east,font=\scriptsize]
    at (6.28,1.63) {0 means correct};


\fill[heatred!32]
    (6.48,3.55) rectangle (7.96,4.21);
\fill[heatred!62]
    (8.05,3.55) rectangle (9.53,4.21);
\fill[heatred!8]
    (9.62,3.55) rectangle (11.10,4.21);

\fill[heatred!59]
    (6.48,2.80) rectangle (7.96,3.46);
\fill[heatred!54]
    (8.05,2.80) rectangle (9.53,3.46);
\fill[heatred!8]
    (9.62,2.80) rectangle (11.10,3.46);

\fill[heatred!61]
    (6.48,2.05) rectangle (7.96,2.71);
\fill[heatred!48]
    (8.05,2.05) rectangle (9.53,2.71);
\fill[heatred!6]
    (9.62,2.05) rectangle (11.10,2.71);

\fill[heatred!7]
    (6.48,1.30) rectangle (7.96,1.96);
\fill[heatred!9]
    (8.05,1.30) rectangle (9.53,1.96);
\fill[heatred!3]
    (9.62,1.30) rectangle (11.10,1.96);


\shade[
    left color=white,
    right color=heatred
]
    (6.48,0.95) rectangle (11.10,1.09);

\node[
    anchor=north,
    font=\scriptsize
] at (6.48,0.87) {0};

\node[
    anchor=north,
    font=\scriptsize
] at (11.10,0.87) {0.5};


\node[
    anchor=north,
    font=\footnotesize,
    text=black!75
] at (1.59,0.48)
{Mean AUROC drop};

\node[
    anchor=north,
    font=\footnotesize,
    text=black!75
] at (8.79,0.48)
{AUROC drop};

\end{tikzpicture}%
}

\caption{
Held-out conditions can substantially degrade the model's verdict and the
readout while leaving the isolated judgment comparatively stable.
Panel (a) shows the mean AUROC drop across the four held-out conditions.
Panel (b) shows the drop separately for each condition.
}
\label{fig:degradation}
\end{figure}

Across the negated question and the three reversed response mappings, AUROC drops by
0.127 for the verdict and 0.129 for the readout, but only 0.024 for the isolated
judgment (Figure~\ref{fig:degradation}). The verdict and readout also fail on
different model--dataset cells: the readout drops where the model applies the
inverted mapping, whereas the verdict drops where it does not
(Appendix~\ref{app:reversed}). In both cases, the isolated judgment remains within
0.03 AUROC of the condition-fitted readout.

\subsection{The Isolated Judgment Remains Readable When Verdicts Disagree}

We test whether the isolated judgment stays stable when negation changes the model's
verdict. We fit the readout to the model's own Yes/No verdicts, then evaluate the
isolated readout on the same inputs under the negated question. If the isolated
judgment changes with the verdict, it should track the negated verdict. If it remains
stable, it should continue to track the same judgment across the instruction change.

Across cells, the isolated judgment predicts the Yes/No verdict at a median of
0.924 AUROC, while it predicts the verdict under the negated question at
0.16--0.86 by cell (median 0.675), with the Yes/No value higher in every
model--dataset cell. We then restrict the comparison to examples where the two
verdicts disagree. Even on these examples, the isolated judgment predicts the
Yes/No verdict at 0.953 AUROC. Thus, even when the model changes its verdict, the
same judgment remains readable internally (Appendix~\ref{app:labelfree}).

\section{The Judgment Is Readable Before Expression Disrupts the Readout}
\label{sec:depth}

The judgment is already readable before changes in how it must be expressed begin
to disrupt the readout. At intermediate layers, the readout is nearly identical
across the Yes/No condition, the negated question, and the condition in which No
means correct, but these conditions diverge later in the network. Importantly, the isolated judgment remains closely aligned across these conditions
throughout model depth. The directions responsible for the late
disruption are tied to changes in response meaning rather than changes in response
symbol.

\subsection{The Expression Subspace Disrupts the Readout Late}

At half of model depth, the readout is nearly identical under Yes/No, No means
correct, and the negated question, reaching 0.786, 0.781, and 0.783 AUROC,
respectively. The three conditions begin to diverge by three quarters of model
depth. By 0.9 of model depth, the readout remains at 0.813 under Yes/No but falls
to 0.648 when No means correct and 0.632 under negation. Thus, the judgment is
already readable before changes in how it must be expressed begin to disrupt the
readout.

\begin{figure}[t]
\centering

\begin{tikzpicture}

\begin{groupplot}[
    group style={
        group size=2 by 1,
        horizontal sep=0.075\linewidth,
        yticklabels at=edge left
    },
    width=0.445\linewidth,
    height=0.335\linewidth,
    xmin=0.23,
    xmax=1.02,
    ymin=0.59,
    ymax=0.87,
    xtick={0.25,0.50,0.75,0.90,1.00},
    xticklabels={.25,.50,.75,.90,1},
    ytick={0.60,0.70,0.80},
    xlabel={Fraction of model depth},
    tick label style={font=\small},
    label style={font=\small},
    title style={font=\small\bfseries, yshift=3pt},
    axis x line*=bottom,
    axis y line*=left,
    axis line style={black!60, line width=0.6pt},
    tick style={black!60, line width=0.6pt},
    ymajorgrids=true,
    grid style={black!10, line width=0.4pt},
]

\nextgroupplot[
    title={(a) Readout}
]

\addplot[
    color={rgb,255:red,70;green,70;blue,70},
    line width=1.15pt,
    mark=o,
    mark options={
        draw={rgb,255:red,70;green,70;blue,70},
        fill=white,
        line width=0.8pt
    }
] coordinates {
    (0.25,0.699)
    (0.50,0.786)
    (0.75,0.818)
    (0.90,0.813)
    (1.00,0.815)
};

\addplot[
    color={rgb,255:red,76;green,120;blue,168},
    line width=1.15pt,
    mark=square*,
    mark options={
        fill={rgb,255:red,76;green,120;blue,168}
    }
] coordinates {
    (0.25,0.700)
    (0.50,0.781)
    (0.75,0.756)
    (0.90,0.648)
    (1.00,0.640)
};

\addplot[
    color={rgb,255:red,225;green,87;blue,89},
    line width=1.15pt,
    dashed,
    mark=triangle*,
    mark options={
        fill={rgb,255:red,225;green,87;blue,89}
    }
] coordinates {
    (0.25,0.699)
    (0.50,0.783)
    (0.75,0.756)
    (0.90,0.632)
    (1.00,0.650)
};

\nextgroupplot[
    title={(b) Isolated judgment}
]

\addplot[
    color={rgb,255:red,70;green,70;blue,70},
    line width=1.15pt,
    mark=o,
    mark options={
        draw={rgb,255:red,70;green,70;blue,70},
        fill=white,
        line width=0.8pt
    }
] coordinates {
    (0.25,0.702)
    (0.50,0.788)
    (0.75,0.824)
    (0.90,0.820)
    (1.00,0.818)
};

\addplot[
    color={rgb,255:red,76;green,120;blue,168},
    line width=1.15pt,
    mark=square*,
    mark options={
        fill={rgb,255:red,76;green,120;blue,168}
    }
] coordinates {
    (0.25,0.701)
    (0.50,0.782)
    (0.75,0.808)
    (0.90,0.799)
    (1.00,0.781)
};

\addplot[
    color={rgb,255:red,225;green,87;blue,89},
    line width=1.15pt,
    dashed,
    mark=triangle*,
    mark options={
        fill={rgb,255:red,225;green,87;blue,89}
    }
] coordinates {
    (0.25,0.702)
    (0.50,0.784)
    (0.75,0.818)
    (0.90,0.811)
    (1.00,0.802)
};

\end{groupplot}

\begin{scope}[overlay]
\node[
    rotate=90,
    font=\small
] at ($(group c1r1.west)+(-0.72cm,0)$)
{AUROC};
\end{scope}

\node[
    anchor=north,
    font=\small
] at
($(group c1r1.south)!0.5!(group c2r1.south)+(0,-0.92cm)$)
{
\begin{tabular}{@{}c@{\hspace{1.4em}}c@{\hspace{1.4em}}c@{}}

\tikz[baseline=-0.45ex]{
    \draw[
        color={rgb,255:red,70;green,70;blue,70},
        line width=1.15pt
    ] (-0.22,0) -- (0.22,0);

    \draw[
        color={rgb,255:red,70;green,70;blue,70},
        fill=white,
        line width=0.8pt
    ] (0,0) circle[radius=1.5pt];
}
\hspace{0.3em} Yes/No

&

\tikz[baseline=-0.45ex]{
    \draw[
        color={rgb,255:red,76;green,120;blue,168},
        line width=1.15pt
    ] (-0.22,0) -- (0.22,0);

    \fill[
        color={rgb,255:red,76;green,120;blue,168}
    ] (-1.5pt,-1.5pt) rectangle (1.5pt,1.5pt);
}
\hspace{0.3em} No means correct

&

\tikz[baseline=-0.45ex]{
    \draw[
        color={rgb,255:red,225;green,87;blue,89},
        line width=1.15pt,
        dashed
    ] (-0.22,0) -- (0.22,0);

    \fill[
        color={rgb,255:red,225;green,87;blue,89}
    ]
    (0,2.1pt)
    -- (-1.8pt,-1.05pt)
    -- (1.8pt,-1.05pt)
    -- cycle;
}
\hspace{0.3em} Negated question

\end{tabular}
};

\path ($(group c1r1.south)+(0,-1.35cm)$);

\end{tikzpicture}

\caption{
Judgment and expression emerge at different depths. At intermediate layers,
the readout is nearly identical across Yes/No, No-means-correct, and the
negated question. The conditions diverge in the later layers (a), while the isolated
judgment remains aligned across depth (b). Full depth-wise results are in
Appendix~\ref{app:depth}.
}
\label{fig:depth}
\end{figure}
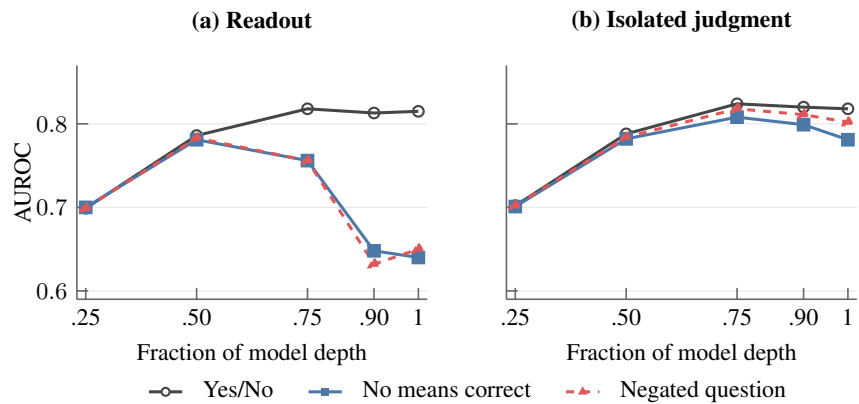

Importantly, the isolated judgment remains closely aligned across all three
conditions throughout model depth. At three quarters of model depth, it reaches
0.818 AUROC under the negated question, compared with 0.802 at the final layer.
The late divergence of the readout is not because the instruction becomes available
only late in the network. When the response form is specified before the judgment
question, condition identity is already decodable at 0.99 AUROC before any response
token is produced. What appears late is specifically the part of the expression
subspace that disrupts the readout. Full depth-wise results are reported in
Appendix~\ref{app:depth}.

\subsection{Response Meaning Drives the Disruption}

The late disruption is associated with changes in what the response means. Changing
among letters, digits, or sentence responses leaves the readout largely unchanged.
The readout instead degrades when ``No'' or ``False'' expresses a positive judgment,
or when the judgment question is negated. The directions associated with these
changes are also sufficient to recover nearly the same performance as the full
expression subspace on the reversed response mappings and the negated question.

The other directions in the expression subspace capture reproducible differences
across response forms, but they contribute little to the readout disruption. The
three directions estimated from the negation pairs recover the negated question
and reversed response mappings to within 0.005 AUROC of the full expression
subspace, whereas the leading direction alone is not sufficient. Thus, the late
disruption is concentrated in a small set of directions associated with changes
in response meaning. Together with the depth-wise results, this shows a clear
separation: the judgment remains readable across depth, while the directions that
interfere with its expression become influential only later in the network
(Appendix~\ref{app:parts}).

\section{Implications and Limitations}
\label{sec:discussion}

\paragraph{Implications.}
A model can retain a readable judgment even when its verdict changes. This changes
what can be inferred from behavioral failure: an incorrect verdict can arise because
the model formed the wrong judgment, or because a judgment that was already present
was expressed incorrectly. A readout from the hidden state does not resolve this
ambiguity on its own, because the readout can also be disrupted by how the judgment
must be expressed. The expression subspace provides a way to separate these two
sources of variation, revealing when a judgment remains readable despite a failure
in the model's final response.

This distinction also changes how model evaluations should be interpreted. Behavioral
accuracy measures only the final verdict, while a readout from the final hidden state
can itself be affected by how the judgment must be expressed. As a result, neither
behavior nor a standard internal readout necessarily reveals whether the judgment
itself is present. Evaluating judgment separately from expression provides a more
direct view of what remains represented internally when the model's final response
fails.

The distinction also has implications for intervention. A failure in judgment and a
failure in expression call for different kinds of correction: the former requires
changing how the model evaluates the input, while the latter may require changing
how an existing judgment is translated into a response. Our results therefore suggest
that locating a failure relative to the judgment can help identify which part of the
computation should be targeted.

\paragraph{Limitations and future work.}
Our results establish a separation between judgment and expression in the hidden
state, not a mechanistic decomposition of the model's computation. Our experiments
focus on binary judgments, single-token or short-sentence verdicts, and models from
7B to 32B parameters. An important next step is to test whether the same separation
extends to open-ended judgments and longer generations, and whether targeted
interventions on judgment and expression can establish distinct causal roles for
the two. Moving from a representational separation to a mechanistic one would
clarify when judgment and expression can be independently read and manipulated.

\subsection*{AI use statement}
Generative AI tools assisted with code drafting, figures, manuscript organization and prose. The authors are responsible for verifying the numerical results, citations, code, and final text.

\subsection*{Ethics statement}
This work studies how language models express correctness, entailment, and safety judgments. The safety experiment uses public, human-labeled BeaverTails examples and generates no new harmful content. No new human-subject data were collected.

\subsection*{Reproducibility statement}
Code, prompts, and per-cell result files are available at an anonymous repository\footnote{\url{https://github.com/jchaemin/isolated-judgment}}. Appendix~\ref{app:protocol} specifies the models, datasets, prompt wordings, folds, benchmark labels, and uncertainty estimates used throughout the paper. The remaining appendices provide the estimation details, controls, and per-cell results underlying the quantitative claims in the main text. All transfer results are evaluated on question-disjoint held-out data.

\bibliographystyle{iclr2027_conference}
\bibliography{iclr2027_conference}

@inproceedings{azaria2023internal,
  title={The internal state of an LLM knows when it’s lying},
  author={Azaria, Amos and Mitchell, Tom},
  booktitle={Findings of the Association for Computational Linguistics: EMNLP 2023},
  pages={967--976},
  year={2023}
}

@inproceedings{bao2025probing,
  title={Probing the geometry of truth: Consistency and generalization of truth directions in LLMs across logical transformations and question answering tasks},
  author={Bao, Yuntai and Zhang, Xuhong and Du, Tianyu and Zhao, Xinkui and Feng, Zhengwen and Peng, Hao and Yin, Jianwei},
  booktitle={Findings of the Association for Computational Linguistics: ACL 2025},
  pages={682--700},
  year={2025}
}

@inproceedings{belrose2023leace,
  title={{LEACE}: Perfect linear concept erasure in closed form},
  author={Belrose, Nora and Schneider-Joseph, David and Ravfogel, Shauli and Cotterell, Ryan and Raff, Edward and Biderman, Stella},
  booktitle={Advances in Neural Information Processing Systems},
  year={2023}
}

@inproceedings{burger2024truth, title={Truth is universal: Robust detection of lies in {LLMs}}, author={B{\"u}rger, Lennart and Hamprecht, Fred A. and Nadler, Boaz}, booktitle={Advances in Neural Information Processing Systems}, year={2024}}

@inproceedings{burns2023latent, title={Discovering latent knowledge in language models without supervision}, author={Burns, Collin and Ye, Haotian and Klein, Dan and Steinhardt, Jacob}, booktitle={International Conference on Learning Representations}, year={2023}}

@article{cobbe2021gsm8k, title={Training verifiers to solve math word problems}, author={Cobbe, Karl and Kosaraju, Vineet and Bavarian, Mohammad and Chen, Mark and Jun, Heewoo and Kaiser, Lukasz and Plappert, Matthias and Tworek, Jerry and Hilton, Jacob and Nakano, Reiichiro and Hesse, Christopher and Schulman, John}, journal={arXiv preprint arXiv:2110.14168}, year={2021}}

@article{farquhar2023challenges, title={Challenges with unsupervised {LLM} knowledge discovery}, author={Farquhar, Sebastian and Varma, Vikrant and Kenton, Zachary and Gasteiger, Johannes and Mikulik, Vladimir and Shah, Rohin}, journal={arXiv preprint arXiv:2312.10029}, year={2023}}

@inproceedings{gekhman2025insideout, title={Inside-out: Hidden factual knowledge in {LLMs}}, author={Gekhman, Zorik and Ben-David, Eyal and Orgad, Hadas and Ofek, Eran and Belinkov, Yonatan and Szpektor, Idan and Herzig, Jonathan and Reichart, Roi}, booktitle={Conference on Language Modeling}, year={2025}}

@article{gemma2024, title={Gemma 2: Improving open language models at a practical size}, author={{Gemma Team} and Riviere, Morgane and Pathak, Shreya and Sessa, Pier Giuseppe and others}, journal={arXiv preprint arXiv:2408.00118}, year={2024}}

@inproceedings{halawi2023overthinking,
  title={Overthinking the truth: Understanding how language models process false demonstrations},
  author={Halawi, Danny and Denain, Jean-Stanislas and Steinhardt, Jacob},
  booktitle={International Conference on Learning Representations},
  volume={2024},
  pages={42749--42787},
  year={2024}
}

@inproceedings{hendrycks2021mmlu, title={Measuring massive multitask language understanding}, author={Hendrycks, Dan and Burns, Collin and Basart, Steven and Zou, Andy and Mazeika, Mantas and Song, Dawn and Steinhardt, Jacob}, booktitle={International Conference on Learning Representations}, year={2021}}

@inproceedings{ji2023beavertails, title={{BeaverTails}: Towards improved safety alignment of {LLM} via a human-preference dataset}, author={Ji, Jiaming and Liu, Mickel and Dai, Josef and Pan, Xuehai and Zhang, Chi and Bian, Ce and Chen, Boyuan and Sun, Ruiyang and Wang, Yizhou and Yang, Yaodong}, booktitle={Advances in Neural Information Processing Systems}, volume={36}, year={2023}}

@inproceedings{joshi2017triviaqa, title={{TriviaQA}: A large scale distantly supervised challenge dataset for reading comprehension}, author={Joshi, Mandar and Choi, Eunsol and Weld, Daniel and Zettlemoyer, Luke}, booktitle={Proceedings of ACL}, pages={1601--1611}, year={2017}}

@article{kadavath2022language, title={Language models (mostly) know what they know}, author={Kadavath, Saurav and Conerly, Tom and Askell, Amanda and Henighan, Tom and Drain, Dawn and Perez, Ethan and Schiefer, Nicholas and Hatfield-Dodds, Zac and DasSarma, Nova and Tran-Johnson, Eli and others}, journal={arXiv preprint arXiv:2207.05221}, year={2022}}

@article{levinstein2024still, title={Still no lie detector for language models: Probing empirical and conceptual roadblocks}, author={Levinstein, B. A. and Herrmann, Daniel A.}, journal={Philosophical Studies}, volume={182}, number={7}, pages={1539--1565}, year={2025}}

@inproceedings{li2023inference, title={Inference-time intervention: Eliciting truthful answers from a language model}, author={Li, Kenneth and Patel, Oam and Vi{\'e}gas, Fernanda and Pfister, Hanspeter and Wattenberg, Martin}, booktitle={Advances in Neural Information Processing Systems}, year={2023}}

@article{lieberum2023does, title={Does circuit analysis interpretability scale? Evidence from multiple choice capabilities in {Chinchilla}}, author={Lieberum, Tom and Rahtz, Matthew and Kram{\'a}r, J{\'a}nos and Nanda, Neel and Irving, Geoffrey and Shah, Rohin and Mikulik, Vladimir}, journal={arXiv preprint arXiv:2307.09458}, year={2023}}

@inproceedings{liu2024hyperplane,
  title={On the universal truthfulness hyperplane inside LLMs},
  author={Liu, Junteng and Chen, Shiqi and Cheng, Yu and He, Junxian},
  booktitle={Proceedings of the 2024 Conference on Empirical Methods in Natural Language Processing},
  pages={18199--18224},
  year={2024}
}

@article{llama3herd2024, title={The {Llama 3} herd of models}, author={Grattafiori, Aaron and Dubey, Abhimanyu and Jauhri, Abhinav and Pandey, Abhinav and Kadian, Abhishek and Al-Dahle, Ahmad and Letman, Aiesha and Mathur, Akhil and others}, journal={arXiv preprint arXiv:2407.21783}, year={2024}}

@inproceedings{marks2024geometry, title={The geometry of truth: Emergent linear structure in large language model representations of true/false datasets}, author={Marks, Samuel and Tegmark, Max}, booktitle={First Conference on Language Modeling}, year={2024}}

@article{mistral7b2023, title={Mistral 7{B}}, author={Jiang, Albert Q. and Sablayrolles, Alexandre and Mensch, Arthur and Bamford, Chris and Chaplot, Devendra Singh and de las Casas, Diego and Bressand, Florian and Lengyel, Gianna and Lample, Guillaume and Saulnier, Lucile and others}, journal={arXiv preprint arXiv:2310.06825}, year={2023}}

@inproceedings{nie2020anli, title={Adversarial {NLI}: A new benchmark for natural language understanding}, author={Nie, Yixin and Williams, Adina and Dinan, Emily and Bansal, Mohit and Weston, Jason and Kiela, Douwe}, booktitle={Proceedings of ACL}, pages={4885--4901}, year={2020}}

@inproceedings{orgad2025know,
  title={Llms know more than they show: On the intrinsic representation of llm hallucinations},
  author={Orgad, Hadas and Toker, Michael and Gekhman, Zorik and Reichart, Roi and Szpektor, Idan and Kotek, Hadas and Belinkov, Yonatan},
  booktitle={International Conference on Learning Representations},
  volume={2025},
  pages={66880--66913},
  year={2025}
}

@article{poulis2026limits, title={Testing the limits of truth directions in {LLMs}}, author={Poulis, Angelos and Crovella, Mark and Terzi, Evimaria}, journal={arXiv preprint arXiv:2604.03754}, year={2026}}

@article{qwen2024technical, title={Qwen2.5 technical report}, author={Yang, An and Yang, Baosong and Zhang, Beichen and Hui, Binyuan and Zheng, Bo and Yu, Bowen and others}, journal={arXiv preprint arXiv:2412.15115}, year={2024}}

@inproceedings{ravfogel2020inlp,
  title={Null it out: Guarding protected attributes by iterative nullspace projection},
  author={Ravfogel, Shauli and Elazar, Yanai and Gonen, Hila and Twiton, Michael and Goldberg, Yoav},
  booktitle={Proceedings of the 58th Annual Meeting of the Association for Computational Linguistics},
  pages={7237--7256},
  year={2020}
}

@article{robinson2023leveraging, title={Leveraging large language models for multiple choice question answering}, author={Robinson, Joshua and Wingate, David}, journal={International Conference on Learning Representations}, year={2023}}

@inproceedings{sclar2024quantifying,
  title={Quantifying Language Models' Sensitivity to Spurious Features in Prompt Design or: How I learned to start worrying about prompt formatting},
  author={Sclar, Melanie and Choi, Yejin and Tsvetkov, Yulia and Suhr, Alane},
  booktitle={International Conference on Learning Representations},
  volume={2024},
  pages={25055--25083},
  year={2024}
}

@inproceedings{servedio2025hidden,
  title={Are the hidden states hiding something? testing the limits of factuality-encoding capabilities in llms},
  author={Servedio, Giovanni and De Bellis, Alessandro and Di Palma, Dario and Anelli, Vito Walter and Di Noia, Tommaso},
  booktitle={Proceedings of the 63rd Annual Meeting of the Association for Computational Linguistics (Volume 1: Long Papers)},
  pages={6089--6104},
  year={2025}
}

@inproceedings{wendler2024llamas,
  title={Do llamas work in english? on the latent language of multilingual transformers},
  author={Wendler, Chris and Veselovsky, Veniamin and Monea, Giovanni and West, Robert},
  booktitle={Proceedings of the 62nd Annual Meeting of the Association for Computational Linguistics (Volume 1: Long Papers)},
  pages={15366--15394},
  year={2024}
}

@inproceedings{wiegreffe2025answer,
  title={Answer, assemble, ace: Understanding how LMs answer multiple choice questions},
  author={Wiegreffe, Sarah and Tafjord, Oyvind and Belinkov, Yonatan and Hajishirzi, Hannaneh and Sabharwal, Ashish},
  booktitle={International Conference on Learning Representations},
  volume={2025},
  pages={78293--78318},
  year={2025}
}

@inproceedings{wong2026bind,
  title={When Models Decide and When They Bind: A Two-Stage Computation for Multiple-Choice Question Answering},
  author={Wong, Hugh Mee and Nouwen, Rick and Gatt, Albert},
  booktitle={Findings of the Association for Computational Linguistics: ACL 2026},
  pages={22804--22821},
  year={2026}
}

@inproceedings{zheng2024large,
  title={Large language models are not robust multiple choice selectors},
  author={Zheng, Chujie and Zhou, Hao and Meng, Fandong and Zhou, Jie and Huang, Minlie},
  booktitle={International Conference on Learning Representations},
  volume={2024},
  pages={19426--19454},
  year={2024}
}

@article{zou2023representation, title={Representation engineering: A top-down approach to {AI} transparency}, author={Zou, Andy and Phan, Long and Chen, Sarah and Campbell, James and Guo, Phillip and Ren, Richard and Pan, Alexander and Yin, Xuwang and Mazeika, Mantas and Dombrowski, Ann-Kathrin and others}, journal={arXiv preprint arXiv:2310.01405}, year={2023}}

@inproceedings{cheang2026recall,
  title={Do {LLMs} Really Know What They Don't Know? Internal States Mainly Reflect Knowledge Recall Rather Than Truthfulness},
  author={Cheang, Chi Seng and Chan, Hou Pong and Zhang, Wenxuan and Deng, Yang},
  booktitle={Findings of the Association for Computational Linguistics: ACL 2026},
  pages={713--730},
  year={2026}
}

@article{elazar2021amnesic,
  title={Amnesic Probing: Behavioral Explanation with Amnesic Counterfactuals},
  author={Elazar, Yanai and Ravfogel, Shauli and Jacovi, Alon and Goldberg, Yoav},
  journal={Transactions of the Association for Computational Linguistics},
  volume={9},
  pages={160--175},
  year={2021}
}

@inproceedings{han2025factuality,
  title={Simple Factuality Probes Detect Hallucinations in Long-Form Natural Language Generation},
  author={Han, Jiatong and Band, Neil and Razzak, Muhammed and Kossen, Jannik and Rudner, Tim G. J. and Gal, Yarin},
  booktitle={Findings of the Association for Computational Linguistics: EMNLP 2025},
  pages={16209--16226},
  year={2025}
}

@inproceedings{hewitt2019designing,
  title={Designing and Interpreting Probes with Control Tasks},
  author={Hewitt, John and Liang, Percy},
  booktitle={Proceedings of the 2019 Conference on Empirical Methods in Natural Language Processing and the 9th International Joint Conference on Natural Language Processing},
  pages={2733--2743},
  year={2019}
}

@inproceedings{lu2022fantastically,
  title={Fantastically Ordered Prompts and Where to Find Them: Overcoming Few-Shot Prompt Order Sensitivity},
  author={Lu, Yao and Bartolo, Max and Moore, Alastair and Riedel, Sebastian and Stenetorp, Pontus},
  booktitle={Proceedings of the 60th Annual Meeting of the Association for Computational Linguistics},
  pages={8086--8098},
  year={2022}
}

@inproceedings{min2022rethinking,
  title={Rethinking the Role of Demonstrations: What Makes In-Context Learning Work?},
  author={Min, Sewon and Lyu, Xinxi and Holtzman, Ari and Artetxe, Mikel and Lewis, Mike and Hajishirzi, Hannaneh and Zettlemoyer, Luke},
  booktitle={Proceedings of the 2022 Conference on Empirical Methods in Natural Language Processing},
  pages={11048--11064},
  year={2022}
}

@inproceedings{park2024linear,
  title={The Linear Representation Hypothesis and the Geometry of Large Language Models},
  author={Park, Kiho and Choe, Yo Joong and Veitch, Victor},
  booktitle={Proceedings of the 41st International Conference on Machine Learning},
  volume={235},
  pages={39643--39666},
  year={2024}
}

@inproceedings{stolfo2025instruction,
  title={Improving instruction-following in language models through activation steering},
  author={Stolfo, Alessandro and Balachandran, Vidhisha and Yousefi, Safoora and Horvitz, Eric and Nushi, Besmira},
  booktitle={International Conference on Learning Representations},
  volume={2025},
  pages={55790--55823},
  year={2025}
}

@inproceedings{subramani2022steering,
  title={Extracting Latent Steering Vectors from Pretrained Language Models},
  author={Subramani, Nishant and Suresh, Nivedita and Peters, Matthew E},
  booktitle={Findings of the Association for Computational Linguistics: ACL 2022},
  pages={566--581},
  year={2022}
}

@inproceedings{webson2022prompts,
  title={Do Prompt-Based Models Really Understand the Meaning of Their Prompts?},
  author={Webson, Albert and Pavlick, Ellie},
  booktitle={Proceedings of the 2022 Conference of the North American Chapter of the Association for Computational Linguistics: Human Language Technologies},
  pages={2300--2344},
  year={2022}
}

@inproceedings{wei2023symbol,
  title={Symbol Tuning Improves In-Context Learning in Language Models},
  author={Wei, Jerry and Hou, Le and Lampinen, Andrew and Chen, Xiangning and Huang, Da and Tay, Yi and Chen, Xinyun and Lu, Yifeng and Zhou, Denny and Ma, Tengyu and Le, Quoc V.},
  booktitle={Proceedings of the 2023 Conference on Empirical Methods in Natural Language Processing},
  pages={968--979},
  year={2023}
}

@inproceedings{zhao2021calibrate,
  title={Calibrate Before Use: Improving Few-Shot Performance of Language Models},
  author={Zhao, Zihao and Wallace, Eric and Feng, Shi and Klein, Dan and Singh, Sameer},
  booktitle={Proceedings of the 38th International Conference on Machine Learning},
  volume={139},
  pages={12697--12706},
  year={2021}
}

@article{clark2018arc,
  title={Think you have solved question answering? {Try ARC}, the {AI2} reasoning challenge},
  author={Clark, Peter and Cowhey, Isaac and Etzioni, Oren and Khot, Tushar and Sabharwal, Ashish and Schoenick, Carissa and Tafjord, Oyvind},
  journal={arXiv preprint arXiv:1803.05457},
  year={2018}
}

\appendix
\renewcommand{\thesection}{\Alph{section}}

\section{Protocol}
\label{app:protocol}

\paragraph{Models and data.} Qwen2.5-7B-Instruct and Qwen2.5-14B-Instruct \citep{qwen2024technical}, Llama-3.1-8B-Instruct \citep{llama3herd2024}, and Mistral-7B-Instruct-v0.3 \citep{mistral7b2023}; Gemma-2-9B-IT \citep{gemma2024} on all four correctness datasets (ARC and GSM8K descriptive, with fewer than 100 negatives), added after the method was fixed, with its own candidate answers and folds built the same way. For correctness, each model's own greedy answer on GSM8K \citep{cobbe2021GSM8K}, MMLU \citep{hendrycks2021MMLU}, ARC-Challenge \citep{clark2018ARC}, and TriviaQA \citep{joshi2017TriviaQA} is used as the candidate answer; the model sees the question and candidate answer and produces a one-token verdict. Each model--dataset cell contains approximately 800 questions. Cells with fewer than 100 negative examples are descriptive; this marks the Qwen2.5-7B and Qwen2.5-14B ARC cells as descriptive and leaves fourteen primary cells. The rule is applied on the full cell; the five-wording analyses use the first 400 questions of each cell. Entailment uses 1,000 ANLI \citep{nie2020anli} examples restricted to entailment and contradiction and balanced; safety uses 1,000 balanced examples from the human-labeled BeaverTails \citep{ji2023beavertails} test split with safe as the positive class.

\paragraph{Conditions and wordings.} The response formats are Yes/No, True/False, A/B, B/A, 1/0, Correct/Incorrect, and the two-sentence format (``The proposed answer is correct.'' / ``The proposed answer is incorrect.'', sharing the first four tokens in all four models; the state is read before the first response token). We additionally use the negated question and three held-out conditions in which No, False, or 0 means correct; their states were extracted after every subspace and readout had been fixed. Each condition uses the same ten instruction wordings, matched one to one and differing only in the clause specifying what each response means. In the control where the response format is specified first, that clause is moved before the question; the rows, benchmark labels, folds, and wordings are otherwise unchanged. The five-wording analyses use the first, third, fifth, seventh, and ninth wordings. The prompt-position readouts of \cref{app:position} and the regularization path of \cref{app:transfer} use all ten; the steering runs use the same five on the first 300 questions of each cell; Qwen2.5-32B and Gemma-2-27B use three (the first, fifth, and ninth) on all rows.

\paragraph{Reads, folds, uncertainty.} Five outer folds grouped by question; standardization estimated on each training fold and carried frozen to held-out states; the class-mean readout is the standardized positive-minus-negative mean difference; AUROC is computed only on held-out questions. Within-cell intervals use 10,000 question-clustered bootstrap resamples shared across compared score vectors; aggregate intervals resample cells. The models are not independent units; per-model values are reported wherever an aggregate is. Main-text tables report means over cells; control tables report medians, so the same readout appears as 0.650 (mean) and 0.694 (median) under the negated question.

\section{The readout}
\label{app:judgment}

\subsection{Transfer across response formats}
\label{app:transfer}
\Cref{tab:a1} gives the per-cell transfer values across response formats.

\paragraph{Regularization path.} The Yes/No logistic probe refitted on a fixed grid with the same frozen states and folds, held-out AUROC: $C=10^{-6}$: source 0.82, frozen A/B 0.81, B/A 0.80; $10^{-5}$: 0.83, 0.81, 0.81; $10^{-4}$: 0.84, 0.82, 0.81; $10^{-3}$: 0.84, 0.79, 0.75; $10^{-2}$: 0.82, 0.73, 0.66; $10^{-1}$: 0.80, 0.71, 0.63; $1$: 0.79, 0.69, 0.62 (ten wordings). As $C \to 0$ the coefficient approaches the class-mean direction; source accuracy peaks at $10^{-3}$ while transfer falls from $10^{-4}$ on.

\begin{table}[htbp]
\centering
\scriptsize
\caption{Per-cell transfer of the frozen Yes/No readout (five wordings): its Yes/No AUROC, its AUROC on A/B, B/A, and sentence states, and the AUROC of a readout fitted on each of those formats.}
\label{tab:a1}
\resizebox{\linewidth}{!}{%
\begin{tabular}{lccccccc}
\toprule
cell & Yes/No own & frozen$\to$A/B & A/B fitted & frozen$\to$B/A & B/A fitted & frozen$\to$sentence & sentence fitted \\
\midrule
Llama-3.1-8B ARC & 0.737 & 0.703 & 0.695 & 0.714 & 0.702 & 0.730 & 0.731 \\
Llama-3.1-8B GSM8K & 0.966 & 0.971 & 0.970 & 0.972 & 0.972 & 0.952 & 0.961 \\
Llama-3.1-8B MMLU & 0.719 & 0.679 & 0.672 & 0.688 & 0.678 & 0.708 & 0.699 \\
Llama-3.1-8B TriviaQA & 0.842 & 0.846 & 0.843 & 0.845 & 0.842 & 0.844 & 0.846 \\
Mistral-7B ARC & 0.730 & 0.662 & 0.696 & 0.649 & 0.670 & 0.685 & 0.690 \\
Mistral-7B GSM8K & 0.826 & 0.814 & 0.822 & 0.812 & 0.821 & 0.829 & 0.827 \\
Mistral-7B MMLU & 0.721 & 0.661 & 0.696 & 0.655 & 0.683 & 0.722 & 0.731 \\
Mistral-7B TriviaQA & 0.815 & 0.796 & 0.808 & 0.792 & 0.809 & 0.810 & 0.806 \\
Qwen2.5-14B GSM8K & 0.937 & 0.939 & 0.938 & 0.938 & 0.938 & 0.934 & 0.933 \\
Qwen2.5-14B MMLU & 0.743 & 0.732 & 0.744 & 0.717 & 0.741 & 0.747 & 0.759 \\
Qwen2.5-14B TriviaQA & 0.859 & 0.862 & 0.863 & 0.838 & 0.866 & 0.862 & 0.862 \\
Qwen2.5-7B GSM8K & 0.896 & 0.887 & 0.887 & 0.873 & 0.882 & 0.853 & 0.859 \\
Qwen2.5-7B MMLU & 0.779 & 0.774 & 0.776 & 0.767 & 0.764 & 0.795 & 0.797 \\
Qwen2.5-7B TriviaQA & 0.840 & 0.842 & 0.843 & 0.835 & 0.839 & 0.833 & 0.827 \\
\bottomrule
\end{tabular}}
\end{table}

\subsection{Readout by prompt position}
\label{app:position}
\paragraph{Before and after the judgment question.} PRE is the final-layer state at the candidate answer's last token, before any correctness request; STEM the state after the judgment question and before the response format is specified; POST the response-position state. Class-mean readouts fitted independently at each give 0.741, 0.812, and 0.818, with POST above PRE in all fourteen cells (paired gain 0.078, question-clustered interval [0.053, 0.104]). The PRE direction and scaler carried unchanged give 0.741, 0.703, 0.730. Ten wordings.

\Cref{tab:a1b} gives the values when the response format is specified first; over the ten primary cells of the three models that follow these prompts, the means are 0.808 at the stem and 0.823 at the response position.

\begin{table}[htbp]
\centering
\small
\caption{Response format specified first, per cell: Yes/No AUROC of the class mean fitted at the end of the judgment question (stem, before any response token) and at the response position. Five wordings. Mistral-7B follows these prompts on few rows and is descriptive. Qwen ARC cells are descriptive.}
\label{tab:a1b}
\begin{tabular}{lcc}
\toprule
cell & stem & response position \\
\midrule
Llama-3.1-8B ARC & 0.691 & 0.746 \\
Llama-3.1-8B GSM8K & 0.967 & 0.971 \\
Llama-3.1-8B MMLU & 0.683 & 0.710 \\
Llama-3.1-8B TriviaQA & 0.810 & 0.827 \\
Mistral-7B ARC & 0.694 & 0.717 \\
Mistral-7B GSM8K & 0.776 & 0.788 \\
Mistral-7B MMLU & 0.712 & 0.705 \\
Mistral-7B TriviaQA & 0.785 & 0.794 \\
Qwen2.5-14B ARC & 0.766 & 0.790 \\
Qwen2.5-14B GSM8K & 0.920 & 0.916 \\
Qwen2.5-14B MMLU & 0.717 & 0.756 \\
Qwen2.5-14B TriviaQA & 0.849 & 0.857 \\
Qwen2.5-7B ARC & 0.711 & 0.781 \\
Qwen2.5-7B GSM8K & 0.865 & 0.830 \\
Qwen2.5-7B MMLU & 0.740 & 0.774 \\
Qwen2.5-7B TriviaQA & 0.833 & 0.845 \\
\bottomrule
\end{tabular}
\end{table}

\subsection{Steering}
\label{app:steering}
The Yes/No class-mean direction is fitted at the residual site at 0.75 of depth and added there at doses $-1$, $-0.5$, $-0.25$, $0$, $0.25$, $0.5$, and $1$ in units of the training states' standard deviation along the direction; direction, sign, and scale are fixed from the Yes/No condition. The logit difference between the response that means correct and the response that means incorrect under each condition changes by 2.3895 logits per unit dose under Yes/No, 1.2955 under A/B, and 0.4553 under B/A, positive in 14/14, 14/14, and 13/14 primary cells; on rows that answer in the requested format at every dose the slopes are 2.388, 1.311, and 0.475. Directions fitted to shuffled benchmark labels give 0.0018, 0.0014, 0.0050 and matched random directions 0.0001, 0.0003, 0.0001. From dose $-1$ to $+1$ the rate of ``correct'' verdicts rises by 9.3, 4.3, and 2.7 percentage points. At 0.5 of depth the slopes are 0.61 (A/B) and 0.49 (B/A); at 0.9 of depth the B/A slope changes sign ($-0.16$), so the effect is a property of the intermediate layers. Five wordings.

\section{The expression subspace}
\label{app:subspace}

\subsection{Estimation}
\label{app:estimation}
\paragraph{Estimation.} For each cell, fold, and wording, the same held-out candidate answers are judged under Yes/No, True/False, A/B, B/A, and the sentence format. On training questions only, we form every paired difference between the five response formats, and the paired difference between the affirmative and negated questions on the entailment and safety inputs of the same model, center each set, and take the leading right singular vectors of the stacked, centered differences. The rank is set per cell, fold, and wording by the split-half rule (the largest rank for which the leading subspaces estimated on two disjoint halves have every principal angle with $|\cos| \ge 0.95$), applied jointly to the correctness and cross-task blocks; the cell medians range from 6 to 9 and their median is 8. The subspace is re-estimated inside every fold. Under format-first prompts, condition identity is linearly decodable at 0.91 AUROC at the end of the candidate answer and 0.99 after the instruction, before any response token (means over the fourteen primary cells). Using only paired differences among the five response formats, the rule gives 6.4 dimensions; with the negation pairs only, 2.9. \Cref{tab:a2} gives the per-cell norm share (standardized coordinates) and the AUROC of the readout and the isolated judgment on B/A and negated-question states.

\begin{table}[htbp]
\centering
\scriptsize
\caption{The expression subspace per cell (paired differences among the five response formats plus entailment and safety negation pairs; rank by the split-half rule, median over folds and wordings): fraction of the Yes/No readout direction's squared norm inside it (standardized coordinates), and the AUROC of the readout and of the isolated judgment on B/A states and on negated-question states. Five wordings.}
\label{tab:a2}
\resizebox{\linewidth}{!}{%
\begin{tabular}{lcccccc}
\toprule
cell & rank & direction norm in subspace & readout, B/A & isolated, B/A & readout, negated & isolated, negated \\
\midrule
Llama-3.1-8B ARC & 7 & 0.524 & 0.714 & 0.729 & 0.718 & 0.732 \\
Llama-3.1-8B GSM8K & 9 & 0.473 & 0.972 & 0.972 & 0.944 & 0.965 \\
Llama-3.1-8B MMLU & 6 & 0.277 & 0.688 & 0.692 & 0.664 & 0.703 \\
Llama-3.1-8B TriviaQA & 8 & 0.504 & 0.845 & 0.846 & 0.726 & 0.841 \\
Mistral-7B ARC & 8 & 0.569 & 0.649 & 0.695 & 0.646 & 0.687 \\
Mistral-7B GSM8K & 6 & 0.444 & 0.812 & 0.823 & 0.744 & 0.817 \\
Mistral-7B MMLU & 6 & 0.444 & 0.655 & 0.693 & 0.705 & 0.737 \\
Mistral-7B TriviaQA & 8 & 0.576 & 0.792 & 0.797 & 0.623 & 0.794 \\
Qwen2.5-14B GSM8K & 8 & 0.355 & 0.938 & 0.939 & 0.818 & 0.939 \\
Qwen2.5-14B MMLU & 8 & 0.544 & 0.717 & 0.726 & 0.441 & 0.712 \\
Qwen2.5-14B TriviaQA & 9 & 0.649 & 0.838 & 0.860 & 0.174 & 0.829 \\
Qwen2.5-7B GSM8K & 7 & 0.510 & 0.873 & 0.879 & 0.478 & 0.866 \\
Qwen2.5-7B MMLU & 7 & 0.660 & 0.767 & 0.779 & 0.733 & 0.767 \\
Qwen2.5-7B TriviaQA & 8 & 0.575 & 0.835 & 0.837 & 0.684 & 0.835 \\
\bottomrule
\end{tabular}}
\end{table}

\subsection{Held-out conditions}
\label{app:heldout}
\Cref{tab:heldout} is the main comparison on the four held-out conditions; \cref{tab:m1full} gives every readout on every target and \cref{tab:m1gemma} the prospective fifth model. \paragraph{Held-out conditions.} The median gap between the isolated judgment and the condition-fitted class mean is within 0.02 on every held-out condition and every wording.

\begin{table}[htbp]
\centering
\small
\caption{Held-out conditions. AUROC against benchmark labels, means over fourteen cells, five wordings. The expression subspace is estimated from the five response formats and the entailment and safety negation pairs; none of these conditions enters it. Verdict is the model's own response margin; readout is the frozen Yes/No readout; pooled is a class mean fitted on labeled A/B and B/A states; condition-fitted is a class mean fitted on the condition itself. Cell-bootstrap intervals on the isolated-minus-condition-fitted gap: [$-0.012$, $0.004$] (negated question), [$-0.043$, $-0.008$] (No means correct), [$-0.028$, $-0.007$] (False means correct), [$-0.005$, $0.003$] (0 means correct).}
\label{tab:heldout}
\begin{tabular}{lccccc}
\toprule
condition & verdict & readout & pooled & isolated & condition-fitted \\
\midrule
No means correct & 0.617 & 0.640 & 0.789 & 0.781 & 0.805 \\
False means correct & 0.593 & 0.665 & 0.771 & 0.787 & 0.803 \\
0 means correct & 0.769 & 0.787 & 0.804 & 0.809 & 0.810 \\
negated question & 0.665 & 0.650 & 0.794 & 0.802 & 0.805 \\
\bottomrule
\end{tabular}
\end{table}

\begin{table}[htbp]
\centering
\small
\caption{Every readout on every target, means over fourteen cells, five wordings. The sentence format has no native response margin. Tuned probe: an L2-regularised logistic readout fitted under Yes/No, with the regularization strength chosen on the source folds.}
\label{tab:m1full}
\begin{tabular}{lcccccc}
\toprule
target & verdict & readout & pooled & tuned probe & isolated & condition-fitted \\
\midrule
Yes/No & 0.788 & 0.815 & 0.809 & 0.824 & 0.818 & 0.815 \\
True/False & 0.779 & 0.815 & 0.812 & 0.811 & 0.814 & 0.814 \\
A/B & 0.744 & 0.798 & 0.803 & 0.774 & 0.808 & 0.804 \\
sentence & -- & 0.807 & 0.801 & 0.776 & 0.806 & 0.809 \\
1/0 & 0.794 & 0.815 & 0.806 & 0.784 & 0.810 & 0.814 \\
correct/incorrect & 0.779 & 0.816 & 0.810 & 0.800 & 0.815 & 0.810 \\
B/A & 0.726 & 0.792 & 0.799 & 0.756 & 0.805 & 0.801 \\
negated question & 0.665 & 0.650 & 0.794 & 0.732 & 0.802 & 0.805 \\
0 means correct & 0.769 & 0.787 & 0.804 & 0.769 & 0.809 & 0.810 \\
False means correct & 0.593 & 0.665 & 0.771 & 0.732 & 0.787 & 0.803 \\
No means correct & 0.617 & 0.640 & 0.789 & 0.706 & 0.781 & 0.805 \\
\bottomrule
\end{tabular}
\end{table}

\begin{table}[htbp]
\centering
\small
\caption{Gemma-2-9B-IT on all four correctness datasets, the prospective fifth model, with the expression subspace estimated from paired differences among Gemma's five response formats and from its entailment and safety negation pairs; means over the four cells (ARC and GSM8K descriptive), five wordings. Predictions were written before any Gemma state existed. Tuned probe as in \cref{tab:m1full}.}
\label{tab:m1gemma}
\begin{tabular}{lcccccc}
\toprule
target & verdict & readout & pooled & tuned probe & isolated & condition-fitted \\
\midrule
Yes/No & 0.801 & 0.812 & 0.802 & 0.812 & 0.812 & 0.812 \\
True/False & 0.800 & 0.812 & 0.809 & 0.794 & 0.804 & 0.812 \\
A/B & 0.779 & 0.812 & 0.807 & 0.787 & 0.815 & 0.811 \\
sentence & -- & 0.802 & 0.793 & 0.762 & 0.804 & 0.807 \\
1/0 & 0.809 & 0.804 & 0.805 & 0.789 & 0.810 & 0.813 \\
correct/incorrect & 0.794 & 0.807 & 0.803 & 0.802 & 0.808 & 0.807 \\
B/A & 0.727 & 0.798 & 0.795 & 0.776 & 0.810 & 0.797 \\
negated question & 0.753 & 0.545 & 0.784 & 0.610 & 0.798 & 0.788 \\
0 means correct & 0.803 & 0.778 & 0.798 & 0.774 & 0.806 & 0.809 \\
False means correct & 0.709 & 0.776 & 0.794 & 0.768 & 0.796 & 0.798 \\
No means correct & 0.740 & 0.630 & 0.798 & 0.648 & 0.802 & 0.798 \\
\bottomrule
\end{tabular}
\end{table}

\subsection{Depth}
\label{app:depth}
The Yes/No readout direction is fitted at each stored depth (0.25, 0.5, 0.75, 0.9, and the final layer) on the training folds of that depth and applied frozen to the other conditions; the expression subspace is re-estimated at each depth from the same pairs as \cref{tab:heldout}. \Cref{tab:depth} gives the means over the fourteen primary cells at each depth.

\begin{table}[htbp]
\centering
\small
\caption{Readout by depth: the Yes/No readout direction fitted at each depth and applied frozen, and the isolated judgment with the expression subspace estimated at that depth, under Yes/No, B/A, No-means-correct, and the negated question. Means over the fourteen primary cells, five wordings.}
\label{tab:depth}
\begin{tabular}{lcccccccc}
\toprule
depth & \multicolumn{2}{c}{Yes/No} & \multicolumn{2}{c}{B/A} & \multicolumn{2}{c}{No means correct} & \multicolumn{2}{c}{negated question} \\
 & readout & isolated & readout & isolated & readout & isolated & readout & isolated \\
\midrule
0.25 & 0.699 & 0.702 & 0.699 & 0.700 & 0.700 & 0.701 & 0.699 & 0.702 \\
0.50 & 0.786 & 0.788 & 0.780 & 0.785 & 0.781 & 0.782 & 0.783 & 0.784 \\
0.75 & 0.818 & 0.824 & 0.800 & 0.813 & 0.756 & 0.808 & 0.756 & 0.818 \\
0.90 & 0.813 & 0.820 & 0.794 & 0.808 & 0.648 & 0.799 & 0.632 & 0.811 \\
1.00 & 0.815 & 0.818 & 0.792 & 0.805 & 0.640 & 0.781 & 0.650 & 0.802 \\
\bottomrule
\end{tabular}
\end{table}

\subsection{Negation and response symbols}
\label{app:parts}
\Cref{tab:r2} estimates the subspace from paired differences among the five response formats only, from the entailment and safety negation pairs only, and from both, and evaluates the isolated judgment on the held-out conditions. \Cref{tab:rank} projects off the leading direction from the negation pairs, the three directions from the negation pairs, and the full subspace.

\begin{table}[htbp]
\centering
\small
\caption{Which pairs each held-out condition needs: the isolated judgment with the subspace estimated from paired differences among the five response formats only, from entailment and safety negation pairs only, or from both, against the readout and the condition-fitted readout. Means over fourteen cells, five wordings.}
\label{tab:r2}
\begin{tabular}{lccccc}
\toprule
condition & readout & five formats only & negation pairs only & both & condition-fitted \\
\midrule
No means correct & 0.640 & 0.722 & 0.777 & 0.781 & 0.805 \\
False means correct & 0.665 & 0.754 & 0.784 & 0.787 & 0.803 \\
0 means correct & 0.787 & 0.805 & 0.810 & 0.809 & 0.810 \\
negated question & 0.650 & 0.733 & 0.801 & 0.802 & 0.805 \\
\bottomrule
\end{tabular}
\end{table}

\begin{table}[htbp]
\centering
\small
\caption{How much of the subspace the repair needs: the Yes/No readout direction with the leading direction from the negation pairs (rank 1), the three directions from the negation pairs (rank 3), and the full subspace (rank 8) removed, against the readout and the condition-fitted readout. Medians over fourteen cells, five wordings.}
\label{tab:rank}
\begin{tabular}{lccccc}
\toprule
condition & readout & rank 1 & rank 3 & rank 8 & condition-fitted \\
\midrule
Yes/No & 0.820 & 0.819 & 0.820 & 0.822 & 0.820 \\
True/False & 0.819 & 0.819 & 0.820 & 0.821 & 0.819 \\
A/B & 0.805 & 0.812 & 0.811 & 0.812 & 0.815 \\
sentence & 0.819 & 0.808 & 0.817 & 0.817 & 0.816 \\
1/0 & 0.809 & 0.811 & 0.814 & 0.812 & 0.815 \\
correct/incorrect & 0.821 & 0.813 & 0.818 & 0.820 & 0.816 \\
B/A & 0.802 & 0.811 & 0.812 & 0.810 & 0.815 \\
negated question & 0.694 & 0.792 & 0.806 & 0.805 & 0.811 \\
0 means correct & 0.766 & 0.814 & 0.820 & 0.815 & 0.819 \\
False means correct & 0.675 & 0.743 & 0.785 & 0.783 & 0.809 \\
No means correct & 0.690 & 0.719 & 0.770 & 0.772 & 0.802 \\
\bottomrule
\end{tabular}
\end{table}

\subsection{Controls and robustness}
\label{app:controls}
\paragraph{Verdict-free estimation, erasure, principal components, wordings.} \Cref{tab:n2,tab:n3,tab:p6}. Principal angles between the subspace and the leading principal components of the same rank of the Yes/No states: mean absolute cosine 0.32; the overlap subspace (directions with cosine at least 0.95) is empty in every cell; the readout direction has 0.490 of its squared norm in the subspace (median) and 0.988 in the principal components. Removing the principal components changes the readout under the negated question by $-0.087$ (median), while removing the subspace changes it by $+0.093$. Rows: the readout and verdict scores of \cref{tab:heldout} are computed on the same rows, all held-out rows (the response-token logit difference exists on every row); the own-verdict scores of \cref{tab:selflabel} use the rows on which the model emits a Yes/No verdict at the same wording.

\begin{table}[htbp]
\centering
\small
\caption{Concept erasure of condition identity against the isolation. LEACE and INLP fitted on the paired states of all conditions (condition identity one-hot), then the readout on the erased states; medians over fourteen cells, five wordings.}
\label{tab:n3}
\begin{tabular}{lcccc}
\toprule
condition & readout & after LEACE & after INLP & isolated \\
\midrule
No means correct & 0.690 & 0.689 & 0.694 & 0.772 \\
False means correct & 0.675 & 0.674 & 0.688 & 0.783 \\
0 means correct & 0.766 & 0.760 & 0.743 & 0.815 \\
negated question & 0.694 & 0.695 & 0.689 & 0.805 \\
\bottomrule
\end{tabular}
\end{table}

\begin{table}[htbp]
\centering
\small
\caption{The subspace without verdict information: estimated from pairs on which the model's verdict is identical under both conditions (identical by expressed judgment, so Yes under ``correct?'' and No under ``incorrect?'' count as identical; both prompts must be answered with one of their two response tokens; 44--74\% of negation pairs are kept per model, almost all drops being disagreements); from all pairs with the model's verdict sign regressed out of the differences before the singular decomposition; and as in \cref{tab:heldout}. Isolated judgment, medians over fourteen cells, five wordings.}
\label{tab:n2}
\begin{tabular}{lcccc}
\toprule
condition & readout & identical-verdict pairs & verdict sign regressed out & as in \cref{tab:heldout} \\
\midrule
No means correct & 0.690 & 0.772 & 0.771 & 0.772 \\
False means correct & 0.675 & 0.763 & 0.783 & 0.783 \\
0 means correct & 0.766 & 0.820 & 0.818 & 0.815 \\
negated question & 0.694 & 0.784 & 0.782 & 0.805 \\
\bottomrule
\end{tabular}
\end{table}

\begin{table}[htbp]
\centering
\scriptsize
\caption{Held-out wordings: the subspace estimated on three instruction wordings and applied at the other two, and the reverse, against the subspace estimated at the evaluation wording; isolated judgment, medians over fourteen cells.}
\label{tab:p6}
\begin{tabular}{llcccc}
\toprule
estimation $\to$ evaluation & condition & readout & other wordings & same wording & fitted \\
\midrule
P01,P03,P05 $\to$ P07,P09 & No means correct & 0.680 & 0.745 & 0.760 & 0.804 \\
P01,P03,P05 $\to$ P07,P09 & False means correct & 0.674 & 0.763 & 0.763 & 0.810 \\
P01,P03,P05 $\to$ P07,P09 & 0 means correct & 0.770 & 0.814 & 0.816 & 0.818 \\
P01,P03,P05 $\to$ P07,P09 & negated question & 0.727 & 0.788 & 0.811 & 0.809 \\
P07,P09 $\to$ P01,P03,P05 & No means correct & 0.704 & 0.760 & 0.780 & 0.801 \\
P07,P09 $\to$ P01,P03,P05 & False means correct & 0.676 & 0.764 & 0.795 & 0.808 \\
P07,P09 $\to$ P01,P03,P05 & 0 means correct & 0.763 & 0.818 & 0.813 & 0.819 \\
P07,P09 $\to$ P01,P03,P05 & negated question & 0.674 & 0.763 & 0.802 & 0.811 \\
\bottomrule
\end{tabular}
\end{table}

\begin{table}[htbp]
\centering
\small
\caption{A different instruction family: five new wordings unrelated to the original ten, with the readout and the expression subspace taken from the original family and applied unchanged. AUROC against benchmark labels for the original-family readout on the new family's states, the original-family isolated judgment on the same states, and a readout fitted on the new family; means (medians) over fourteen cells.}
\label{tab:family}
\begin{tabular}{lccc}
\toprule
condition & readout & isolated judgment & fitted on the new family \\
\midrule
Yes/No & 0.810 (0.819) & 0.813 (0.816) & 0.814 (0.820) \\
True/False & 0.810 (0.815) & 0.807 (0.815) & 0.812 (0.818) \\
A/B & 0.795 (0.814) & 0.802 (0.816) & 0.800 (0.813) \\
B/A & 0.778 (0.779) & 0.795 (0.809) & 0.797 (0.812) \\
sentence & 0.806 (0.817) & 0.804 (0.814) & 0.804 (0.815) \\
negated question & 0.652 (0.695) & 0.792 (0.793) & 0.804 (0.801) \\
No means correct & 0.595 (0.617) & 0.768 (0.763) & 0.798 (0.805) \\
\bottomrule
\end{tabular}
\end{table}

\paragraph{Subspace controls.} The Yes/No class mean has 0.507 [0.452, 0.559] of its squared norm in the expression subspace. A random subspace of the same rank leaves the readout unchanged.

\subsection{Entailment and safety as judgment tasks}
\label{app:tasks}
The four primary models judge ANLI (entailment) and BeaverTails (safety) under the eleven conditions, 400 of the 1,000 rows each, five wordings, with the first listed response expressing the positive judgment (entailed, safe). The expression subspace is estimated from paired differences among the task's five response formats and from the correctness affirmative/negated pairs, so the directions identified from negation transfer from correctness into each task; the task's own negated question and its held-out No-means-positive, False-means-positive, and 0-means-positive conditions never enter the estimate. Predictions were written before extraction. Per model, the frozen Yes/No readout is within 0.003--0.010 of condition-fitted on entailment and 0.005--0.035 on safety; the isolated judgment is within 0.008--0.028 (entailment) and 0.009--0.028 (safety) of condition-fitted under the negated question and No-means-positive.

\begin{table}[htbp]
\centering
\scriptsize
\caption{Entailment (ANLI) and safety (BeaverTails) as judgment tasks: AUROC against the task label for the readout (frozen Yes/No readout), the isolated judgment, a readout fitted on the condition, and the model's verdict. Means over the four primary models, five wordings. Conditions below the rule are held out from the subspace.}
\label{tab:tasks}
\begin{tabular}{lcccccccc}
\toprule
 & \multicolumn{4}{c}{entailment} & \multicolumn{4}{c}{safety} \\
condition & readout & isolated & fitted & verdict & readout & isolated & fitted & verdict \\
\midrule
Yes/No & 0.866 & 0.866 & 0.866 & 0.864 & 0.903 & 0.902 & 0.903 & 0.894 \\
True/False & 0.866 & 0.860 & 0.866 & 0.865 & 0.905 & 0.904 & 0.907 & 0.891 \\
A/B & 0.848 & 0.846 & 0.846 & 0.812 & 0.888 & 0.896 & 0.901 & 0.889 \\
B/A & 0.851 & 0.848 & 0.846 & 0.767 & 0.878 & 0.894 & 0.899 & 0.829 \\
sentence & 0.845 & 0.846 & 0.838 & -- & 0.874 & 0.895 & 0.896 & -- \\
\midrule
negated question & 0.504 & 0.827 & 0.827 & 0.627 & 0.480 & 0.891 & 0.881 & 0.782 \\
1/0 & 0.846 & 0.849 & 0.845 & 0.833 & 0.893 & 0.896 & 0.896 & 0.887 \\
word pair & 0.840 & 0.846 & 0.843 & 0.807 & 0.880 & 0.891 & 0.894 & 0.889 \\
0 means positive & 0.812 & 0.852 & 0.840 & 0.738 & 0.803 & 0.893 & 0.894 & 0.869 \\
False means positive & 0.781 & 0.858 & 0.854 & 0.510 & 0.756 & 0.890 & 0.884 & 0.702 \\
No means positive & 0.661 & 0.858 & 0.846 & 0.514 & 0.522 & 0.885 & 0.894 & 0.793 \\
\bottomrule
\end{tabular}
\end{table}

\subsection{Additional models}
\label{app:models}
Every model below was run after the method was fixed, with predictions written before extraction, on the reduced protocol: TriviaQA and MMLU, 400 rows, the model's own candidate answers, conditions Yes/No, True/False, A/B, B/A, sentence, No-means-correct, and the negated question, five wordings (three, on all rows, for the two models of 27B and above), ANLI and BeaverTails negation pairs extracted for the model, subspace by the frozen recipe. Models of 27B and above run in 4-bit (NF4). A cell with fewer than 100 negatives among its 400 rows is marked descriptive. Columns: the Yes/No readout; the largest gap between the frozen Yes/No readout and a condition-fitted readout under A/B, B/A, and True/False; the fall of the readout under the negated question; and the largest gap between the isolated judgment and a condition-fitted readout under the negated question and No-means-correct.

\begin{table}[htbp]
\centering
\scriptsize
\caption{Larger models, means over the two cells (TriviaQA, MMLU); three wordings, all rows, rank fixed at 8.}
\label{tab:large}
\begin{tabular}{lcccc}
\toprule
model & Yes/No readout & frozen vs fitted & fall under negation & isolated vs fitted \\
\midrule
Qwen2.5-32B-Instruct & 0.804 & 0.016 & 0.380 & 0.007 \\
Gemma-2-27B-IT & 0.750 & 0.016 & 0.140 & 0.006 \\
\bottomrule
\end{tabular}
\end{table}

\begin{table}[htbp]
\centering
\scriptsize
\caption{Additional models by cell.}
\label{tab:additional}
\resizebox{\linewidth}{!}{%
\begin{tabular}{lcccccccc}
\toprule
 & \multicolumn{4}{c}{TriviaQA} & \multicolumn{4}{c}{MMLU} \\
model & Yes/No & frozen vs fitted & fall & isolated vs fitted & Yes/No & frozen vs fitted & fall & isolated vs fitted \\
\midrule
Mistral-Nemo-12B-Instruct & 0.776 & 0.007 & 0.151 & 0.002 & 0.756 & 0.007 & 0.033 & 0.002 \\
Ministral-8B-Instruct & 0.807 & 0.007 & 0.141 & 0.007 & 0.788 & 0.018 & 0.062 & 0.008 \\
OLMo-2-7B-Instruct & 0.764 & 0.006 & 0.014 & 0.005 & 0.737 & 0.012 & 0.030 & 0.002 \\
\bottomrule
\end{tabular}}
\end{table}

\subsection{Who wrote the candidate}
\label{app:candidates}
\emph{Another model's answers.} Each primary model judges another model's candidate answers to the same TriviaQA and MMLU questions (Qwen2.5-7B reads Llama-3.1-8B, Llama-3.1-8B reads Mistral-7B, Mistral-7B reads Qwen2.5-14B, Qwen2.5-14B reads Qwen2.5-7B), 400 rows, five wordings. The Yes/No readout fitted on the judge's own answers is applied frozen to these states and compared with a readout fitted on them. Predictions were written before extraction. \emph{Dataset-written candidates.} On MMLU and ARC the candidates are the keyed option and one distractor per question, 200 of each, so no model wrote them and labels are balanced; the readout and subspace are fitted on this set.

\begin{table}[htbp]
\centering
\scriptsize
\caption{Another model's answers: the judge's own-answer readout applied to states where it judges another model's answers, against a readout fitted on those states, under Yes/No. Descriptive: Mistral-7B reads Qwen2.5-14B MMLU (82 negatives).}
\label{tab:crossjudge}
\begin{tabular}{lccc}
\toprule
judge reads candidate & negatives & own-fitted & fitted here \\
\midrule
Qwen2.5-7B reads Llama-3.1-8B, TriviaQA & 115 & 0.801 & 0.800 \\
Qwen2.5-7B reads Llama-3.1-8B, MMLU & 128 & 0.810 & 0.810 \\
Llama-3.1-8B reads Mistral-7B, TriviaQA & 114 & 0.828 & 0.825 \\
Llama-3.1-8B reads Mistral-7B, MMLU & 160 & 0.763 & 0.771 \\
Mistral-7B reads Qwen2.5-14B, TriviaQA & 145 & 0.822 & 0.822 \\
Mistral-7B reads Qwen2.5-14B, MMLU (descr.) & 82 & 0.706 & 0.716 \\
Qwen2.5-14B reads Qwen2.5-7B, TriviaQA & 181 & 0.903 & 0.902 \\
Qwen2.5-14B reads Qwen2.5-7B, MMLU & 116 & 0.842 & 0.843 \\
\bottomrule
\end{tabular}
\end{table}

\begin{table}[htbp]
\centering
\scriptsize
\caption{Dataset-written candidates (MMLU and ARC, keyed option and one distractor, balanced). AUROC against the label, means (medians) over eight cells; readout and subspace fitted on this candidate set.}
\label{tab:options}
\begin{tabular}{lcccc}
\toprule
condition & readout & isolated & fitted & verdict \\
\midrule
Yes/No & 0.885 (0.890) & 0.880 (0.884) & 0.885 (0.890) & 0.881 (0.888) \\
True/False & 0.881 (0.888) & 0.868 (0.870) & 0.881 (0.888) & 0.876 (0.887) \\
A/B & 0.865 (0.883) & 0.863 (0.882) & 0.862 (0.877) & 0.793 (0.800) \\
B/A & 0.848 (0.875) & 0.855 (0.880) & 0.854 (0.873) & 0.753 (0.740) \\
sentence & 0.867 (0.873) & 0.865 (0.877) & 0.875 (0.888) & -- \\
negated question & 0.518 (0.502) & 0.827 (0.849) & 0.854 (0.861) & 0.574 (0.613) \\
No means correct & 0.606 (0.714) & 0.821 (0.825) & 0.856 (0.856) & 0.538 (0.486) \\
\bottomrule
\end{tabular}
\end{table}

\section{The isolated judgment and the model's verdict}
\label{app:isolated}

\subsection{The negated question}
\label{app:negation}
\paragraph{Negation.} \Cref{tab:a3b} gives per-cell values. The directions associated with negation are estimated as in \cref{app:estimation} from the paired differences between affirmative and negated questions on ANLI and BeaverTails for the same model, and removed from the correctness Yes/No readout direction; it never sees correctness states or benchmark labels. The per-condition ablation is in \cref{tab:r2}.

\begin{table}[htbp]
\centering
\scriptsize
\caption{The negated question per cell: AUROC of the model's verdict under ``correct?'' and ``incorrect?''; the readout under each; the isolated judgment under ``incorrect?''; and a readout fitted on the negated states. Five wordings.}
\label{tab:a3b}
\resizebox{\linewidth}{!}{%
\begin{tabular}{lcccccc}
\toprule
cell & verdict, correct? & verdict, incorrect? & readout, correct? & readout, incorrect? & isolated, incorrect? & fitted, incorrect? \\
\midrule
Llama-3.1-8B ARC & 0.698 & 0.528 & 0.737 & 0.718 & 0.732 & 0.727 \\
Llama-3.1-8B GSM8K & 0.852 & 0.821 & 0.966 & 0.944 & 0.965 & 0.968 \\
Llama-3.1-8B MMLU & 0.670 & 0.581 & 0.719 & 0.664 & 0.703 & 0.715 \\
Llama-3.1-8B TriviaQA & 0.832 & 0.798 & 0.842 & 0.726 & 0.841 & 0.839 \\
Mistral-7B ARC & 0.708 & 0.557 & 0.730 & 0.646 & 0.687 & 0.663 \\
Mistral-7B GSM8K & 0.792 & 0.761 & 0.826 & 0.744 & 0.817 & 0.822 \\
Mistral-7B MMLU & 0.693 & 0.603 & 0.721 & 0.705 & 0.737 & 0.717 \\
Mistral-7B TriviaQA & 0.829 & 0.707 & 0.815 & 0.623 & 0.794 & 0.799 \\
Qwen2.5-14B GSM8K & 0.898 & 0.898 & 0.937 & 0.818 & 0.939 & 0.935 \\
Qwen2.5-14B MMLU & 0.731 & 0.689 & 0.743 & 0.441 & 0.712 & 0.735 \\
Qwen2.5-14B TriviaQA & 0.855 & 0.853 & 0.859 & 0.174 & 0.829 & 0.867 \\
Qwen2.5-7B GSM8K & 0.883 & 0.730 & 0.896 & 0.478 & 0.866 & 0.875 \\
Qwen2.5-7B MMLU & 0.757 & 0.351 & 0.779 & 0.733 & 0.767 & 0.767 \\
Qwen2.5-7B TriviaQA & 0.838 & 0.432 & 0.840 & 0.684 & 0.835 & 0.848 \\
\bottomrule
\end{tabular}}
\end{table}

\subsection{When No, False, or 0 means correct}
\label{app:reversed}
Per model, the model's verdict scored against the benchmark label when No, False, or 0 means correct (means over its cells, five wordings): No means correct 0.479 (Llama-3.1-8B), 0.579 (Mistral-7B), 0.665 (Qwen2.5-7B), 0.802 (Qwen2.5-14B); False means correct 0.469, 0.453, 0.787, 0.753; 0 means correct 0.754, 0.711, 0.796, 0.838. Across cells, the readout's drop from its Yes/No value correlates with the model's rate of emitting the token that correctly expresses the benchmark label in the corresponding condition: Spearman 0.62 for the negated question, 0.60 when No means correct, 0.54 when False means correct (three wordings); for the letter and digit conditions there is no relation (the readout does not drop there).

\subsection{Readouts fitted to model verdicts}
\label{app:labelfree}
\paragraph{Readout fitted to model verdicts.} For each cell, fold, and wording, we use the sign of the model's native Yes/No margin on training rows where it answered Yes or No to define the two classes in the same standardized coordinates. The isolated version removes the same expression subspace.

\Cref{tab:selflabel} gives the medians over cells, \cref{tab:vcell} the per-cell values, and \cref{tab:l2a} the negated-question readout against the model's Yes/No verdict and against its negated-prompt verdict. Rows: the readout and verdict scores of \cref{tab:heldout} are computed on the same rows, all held-out rows; the own-verdict scores use the rows on which the model emits a Yes/No verdict at the same wording.

\begin{table}[htbp]
\centering
\small
\caption{Readouts fitted to model verdicts: the class mean whose classes are the model's own Yes/No verdicts, before and after removing the expression subspace, scored by AUROC against the model's own Yes/No verdict on each condition's held-out states. Medians over the cells with a defined verdict, five wordings; in parentheses, the range over cells.}
\label{tab:selflabel}
\begin{tabular}{lcc}
\toprule
condition & readout & isolated judgment \\
\midrule
Yes/No & 0.996 (0.98--1.00) & 0.988 (0.94--1.00) \\
True/False & 0.991 (0.94--1.00) & 0.974 (0.89--1.00) \\
A/B & 0.977 (0.87--1.00) & 0.969 (0.86--0.99) \\
sentence & 0.970 (0.86--0.99) & 0.947 (0.84--0.99) \\
1/0 & 0.988 (0.93--1.00) & 0.970 (0.86--1.00) \\
correct/incorrect & 0.986 (0.90--1.00) & 0.977 (0.89--1.00) \\
B/A & 0.976 (0.85--0.99) & 0.969 (0.85--1.00) \\
negated question & 0.354 (0.09--0.84) & 0.925 (0.71--0.97) \\
0 means correct & 0.917 (0.70--0.99) & 0.963 (0.85--1.00) \\
False means correct & 0.727 (0.07--0.86) & 0.910 (0.81--0.98) \\
No means correct & 0.628 (0.06--0.88) & 0.913 (0.71--0.97) \\
\bottomrule
\end{tabular}
\end{table}

\begin{table}[htbp]
\centering
\scriptsize
\caption{Readouts fitted to model verdicts per cell: AUROC against the model's own Yes/No verdict, for the readout fitted to the model's verdicts (plain) and the same with the expression subspace removed (isolated), under the negated question and two reversed conditions. Five wordings. Thirteen cells; Mistral-7B ARC has no defined verdict class.}
\label{tab:vcell}
\begin{tabular}{lcccccc}
\toprule
cell & \multicolumn{2}{c}{negated} & \multicolumn{2}{c}{No means correct} & \multicolumn{2}{c}{False means correct} \\
 & plain & isolated & plain & isolated & plain & isolated \\
\midrule
Llama-3.1-8B ARC & 0.427 & 0.762 & 0.757 & 0.860 & 0.829 & 0.883 \\
Llama-3.1-8B GSM8K & 0.425 & 0.935 & 0.845 & 0.922 & 0.817 & 0.910 \\
Llama-3.1-8B MMLU & 0.488 & 0.708 & 0.797 & 0.790 & 0.798 & 0.806 \\
Llama-3.1-8B TriviaQA & 0.226 & 0.905 & 0.610 & 0.930 & 0.817 & 0.920 \\
Mistral-7B GSM8K & 0.204 & 0.857 & 0.829 & 0.936 & 0.856 & 0.983 \\
Mistral-7B MMLU & 0.354 & 0.820 & 0.878 & 0.890 & 0.727 & 0.832 \\
Mistral-7B TriviaQA & 0.353 & 0.951 & 0.628 & 0.962 & 0.692 & 0.878 \\
Qwen2.5-14B GSM8K & 0.686 & 0.973 & 0.373 & 0.967 & 0.744 & 0.968 \\
Qwen2.5-14B MMLU & 0.261 & 0.827 & 0.319 & 0.706 & 0.355 & 0.826 \\
Qwen2.5-14B TriviaQA & 0.085 & 0.933 & 0.057 & 0.812 & 0.072 & 0.857 \\
Qwen2.5-7B GSM8K & 0.255 & 0.941 & 0.318 & 0.944 & 0.214 & 0.934 \\
Qwen2.5-7B MMLU & 0.841 & 0.925 & 0.698 & 0.878 & 0.331 & 0.925 \\
Qwen2.5-7B TriviaQA & 0.759 & 0.938 & 0.431 & 0.913 & 0.286 & 0.959 \\
\bottomrule
\end{tabular}
\end{table}

\begin{table}[htbp]
\centering
\small
\caption{Under the negated question, the isolated judgment fitted to the model's own Yes/No verdicts (with the expression subspace removed, no benchmark label) scored against the model's Yes/No verdict and against its negated-prompt verdict, on rows where both exist; ``verdicts agree'' is the fraction of rows on which the two verdicts express the same judgment; the last column restricts to rows on which they disagree. Means over five wordings per cell. --: fewer than twenty disagreeing rows.}
\label{tab:l2a}
\resizebox{\linewidth}{!}{%
\begin{tabular}{lcccc}
\toprule
cell & verdicts agree & vs Yes/No verdict & vs negated verdict & vs Yes/No, disagreeing rows \\
\midrule
Llama-3.1-8B ARC & 0.219 & 0.762 & 0.287 & 0.967 \\
Llama-3.1-8B GSM8K & 0.682 & 0.935 & 0.692 & 0.963 \\
Llama-3.1-8B MMLU & 0.251 & 0.708 & 0.394 & 0.855 \\
Llama-3.1-8B TriviaQA & 0.687 & 0.905 & 0.732 & 0.760 \\
Mistral-7B GSM8K & 0.801 & 0.857 & 0.783 & 0.877 \\
Mistral-7B MMLU & 0.476 & 0.820 & 0.403 & 0.817 \\
Mistral-7B TriviaQA & 0.632 & 0.951 & 0.675 & 0.944 \\
Qwen2.5-14B GSM8K & 0.898 & 0.973 & 0.863 & -- \\
Qwen2.5-14B MMLU & 0.651 & 0.826 & 0.409 & 0.962 \\
Qwen2.5-14B TriviaQA & 0.785 & 0.933 & 0.715 & 0.995 \\
Qwen2.5-7B GSM8K & 0.863 & 0.941 & 0.677 & 0.872 \\
Qwen2.5-7B MMLU & 0.288 & 0.924 & 0.162 & 0.977 \\
Qwen2.5-7B TriviaQA & 0.342 & 0.939 & 0.279 & 0.978 \\
\bottomrule
\end{tabular}}
\end{table}

\end{document}